\documentclass[a4paper,fleqn]{cas-dc}

\usepackage[numbers]{natbib}

\usepackage{bm}
\usepackage{amsmath,amsfonts}
\usepackage{algorithmic}
\usepackage{algorithm}
\usepackage{array}
\usepackage{textcomp}
\usepackage{stfloats}
\usepackage{xurl}
\usepackage{verbatim}
\usepackage{graphicx}
\usepackage{cite}
\usepackage{booktabs}
\usepackage{hyperref}
\usepackage{subcaption}
\usepackage{multirow}
\usepackage{placeins}
\usepackage{siunitx}
\usepackage{longtable}
\usepackage[version=3]{acro}
\usepackage{setspace}
\usepackage[titles]{tocloft}

\usepackage{microtype}

\usepackage[color=red!45]{todonotes}
\setuptodonotes{inline}
\newcommand{\ie}{i.\,e.}
\newcommand{\eg}{e.\,g.}
\robustify\bfseries

\begin{document}

\vspace*{1cm}
\begin{center}
\begin{minipage}[t]{0.8\textwidth}
\begin{Large}
This preprint has not undergone peer review or any post-submission improvements or corrections. The Version of Record of this article is published in Journal of Intelligent Manufacturing, and is available online at \url{https://doi.org/10.1007/s10845-023-02313-y}.
\end{Large}
\end{minipage}
\end{center}
\thispagestyle{empty}
\clearpage
\setcounter{page}{1}

\let\WriteBookmarks\relax
\def\floatpagepagefraction{1}
\def\textpagefraction{.001}
\shorttitle{Image Segmentation for Drilling Tool Wear Detection}
\shortauthors{E Schlager et~al.}

\title [mode = title]{Evaluation of Data Augmentation and Loss Functions in Semantic Image Segmentation for Drilling Tool Wear Detection}    
\tnotemark[1]

\author[1]{Elke Schlager}
\ead{elke@schlager.bz}
\cormark[1]
\cortext[cor1]{Corresponding author}
\address[1]{Know-Center GmbH, Sandgasse 36, 8010 Graz, Austria}

\author[1,2,3,4]{Andreas Windisch}
\ead{awindisch@know-center.at}
\address[2]{Graz University of Technology, Institute of Interactive Systems and Data Science, Sandgasse 36, 8010 Graz, Austria}
\address[3]{Washington University in St. Louis, Physics Department, One Brookings Drive, St. Louis, MO 63130, USA}
\address[4]{Reinforcement Learning Community, AI Austria, Wollzeile 24/12, 1010 Vienna, Austria}

\author[5]{Lukas Hanna}
\ead{lukas.hanna@mcl.at}
\address[5]{Materials Center Leoben Forschung GmbH, Roseggerstrasse 12, 8700 Leoben, Austria}

\author[5]{Thomas Klünsner}
\ead{thomas.kluensner@mcl.at}

\author[5]{Elias Jan Hagendorfer}
\ead{eliasjan.hagendorfer@mcl.at}

\author[6]{Tamara Teppernegg}
\ead{tamara.teppernegg@ceratizit.com}
\address[6]{CERATIZIT Austria GmbH, Metallwerk-Plansee-Straße 71, 6600 Breitenwang, Austria}

\tnotetext[1]{Code available here: \url{https://github.com/eschlager/UNet-Drilling}}

\begin{abstract}
Tool wear monitoring is crucial for quality control and cost reduction in manufacturing processes, of which drilling applications are one example. In this paper, we present a U-Net based semantic image segmentation pipeline, deployed on microscopy images of cutting inserts,  for the purpose of wear detection. The wear area is differentiated in two different types, resulting in a multiclass classification problem. Joining the two wear types in one general wear class, on the other hand, allows the problem to be formulated as a binary classification task. Apart from the comparison of the binary and multiclass problem, also different loss functions, \ie, Cross Entropy, Focal Cross Entropy, and a loss based on the Intersection over Union (IoU), are investigated. Furthermore, models are trained on image tiles of different sizes, and augmentation techniques of varying intensities are deployed. We find, that the best performing models are binary models, trained on data with moderate augmentation and an IoU-based loss function.
\end{abstract}

\begin{keywords}
Semantic Image Segmentation \sep Image Data Augmentation \sep Drilling Tool \sep Wear Segmentation
\end{keywords}

\maketitle

\section{Introduction}

Cutting tool degradation does affect product quality, machine health, and production safety. Cutting tools that are replaced too early, that is, with little wear, result in a waste of material and time. In contrast, replacement that is done after the wear has become substantial could have even more severe consequences, such as damaged work pieces, tool failure, and also poses a hazard to workers and machines. Hence, both, too early and too late replacement result in a waste of material and costs. Therefore, monitoring of cutting tool wear is necessary for optimising machining operations \citep{Colantonio2021, Lin2021, Lutz2019}. Tool wear condition monitoring can either follow an indirect or a direct approach. 
Indirect approaches infer tool wear from signals, such as cutting forces, spindle speed, torque, vibration, or acoustics \citep{Lin2021, Lutz2019, Shurrab2021, Wu2017}. The evolution of wear varies even at steady state cutting parameters, and is machine and material dependent. Thus, indirect approaches are very challenging, and no standard tool for wear monitoring is available yet \citep{Colantonio2021, Bergs2020}.
In contrast, direct approaches usually observe tool wear via optical inspection, \eg, based on images. Direct end-to-end approaches, such as in \citet{Lin2021, Holst2022, Qin2022, Miao2021, Klancnik2015}, are sparse in literature and of limited application, since consistent image capture during the machining process is difficult to accomplish.
In semantic image segmentation pixels are marked with respect to predefined categories, \eg, wear classes. This can be used in direct approaches to flag the wear areas and infer the severity of degradation, but also 
for automated generation of ground truth data needed in indirect approaches.



Apart from machine learning techniques such as decision trees and random forests, also deep learning is already broadly used in indirect tool wear monitoring approaches \citep{Shurrab2021, Wu2017, Sanjay2005, Martinez2019, Klaic2014}. In direct approaches, deployment of deep learning is still widely unexplored \citep{Colantonio2021}, and the works in literature are often based on feature extraction methods relying on classical image vision techniques such as filtering, edge detection and morphological operations \citep{Klancnik2015, Moldovan2017, Zhang2012}.  

Deep learning methods require a large amount of training data in general. Especially in image processing tasks, a large amount of labelled data may not be possible to acquire, or very laborious to create. \citet{Ronneberger2015} proposed the ``U-Net for Image Segmentation'' approach, which works efficiently with a smaller amount of training data as compared to other algorithms that have been used for that purpose before. Initially proposed with regard to biomedical images, U-Net is now used in a variety of fields, including segmentation of wear in cutting tool images.

In the work of \citet{Bergs2020}, a tool classification method and a wear segmentation via U-Net for images of four different tool geometries is proposed, namely ball-end mill, end mill, insert and drill tools. The images are of size $512\times512$ pixels, at different magnification levels, with 50 images available per tool geometry and magnification level, resulting in 400 images in total. After applying data augmentation methods, such as image flipping, zooming, rotating, adding blur, and changing contrast, a total of 3000 data samples is obtained. 
The network was trained for all tool types together, as well as for each tool type separately. 
The average of Intersection over Union (IoU) scores on the test sets for the different tool types range between 0.7 (for ball-end mill) and 0.87 (for drill). 
Based on \citet{Bergs2020}, the work of \citet{Holst2022} establishes a pipeline for tool detection, \ie, whether a tool object is within the field of view of the camera, followed by wear area segmentation, and wear metric calculations $VB_{max}$, $VB_{mean}$, and $VB_{area}$.

\citet{Miao2021} propose a direct approach for wear detection of turning machines. Due to the limited space in the machine, the CCD camera and additional light cannot be positioned optimally, resulting in darkened images with blurred boundaries. The data comprises 186 grey-scale images from 14 different tools, of which 111 are taken as training set. The training data is extended to 1110 images using only the linear transformations horizontal flip and rotation, horizontal and vertical offsets, as well as shear transformation. Thus, no change in contrasts, brightness, or blurriness are made. The images of sizes $1280\times1024$ pixels were resized to $320\times256$ pixels for being processed in the U-Net to reduce computational costs. 
Thereby, the basic U-Net is compared with U-Net architectures enhanced by deep supervision and/or attention gates. Furthermore, the authors compare the models' performances when trained based on three different loss functions: binary cross entropy, a loss based on Intersection over Union, and Matthews correlation coefficient (MCC), which tackles the class imbalance issue. When using deep supervision, the losses at the separate levels of the U-Net are added together. Furthermore, elastic net regularisation (based on the weights of all levels when using deep supervision) is added to the loss. U-Net with deep supervision and MCC-based loss performs best, reaching an average Dice coefficient of 0.936 on the test set across five training repetitions.

\citet{Lin2021} propose an automatised wear detection of spiral end milling cutter without the need of positioning the cutter in a specific way, but using image stitching. For segmentation of the wear area, Segnet, Autoencoder, and U-Net were trained and evaluated, with the U-Net performing best.

However, also convolutional neural networks (CNN) other than U-Net are used for semantic image segmentation in tool wear monitoring:
\citet{Qin2022} perform segmentation on stitched images of end milling cutters, but using a fully convolutional neural network (FCN) \citep{Long2015} based on the VGG16 
 \citep{Simonyan2014} structure as feature extraction network.
\citet{Lutz2019} propose a tool wear assessment procedure on a machine tool with exchangeable inserts. The image segmentation is performed pixel-wise by using a sliding window of fixed size, using a CNN to get a label for each pixel. Each pixel can thereby be classified as background, undamaged tool body, or having wear defects of a specified wear type. Because of the pixel-based approach, with each image having the class label of the centre pixel, the training data can be chosen to be balanced with respect to those classes. Morphological operations are then used for removal of noise and small holes of the predicted image mask. Subsequently, post-processed mask-relevant wear metrics are calculated, \eg, 80th percentile of the flank wear width.

In this paper, we perform semantic image segmentation by means of a U-Net, deployed on high-resolution light optical microscopy images of cutting edges. Since the microscopy images are too large for training and predicting as a whole, they are split into small segments, called tiles. Predictions of the whole images can then be made using the overlap-tile strategy proposed by \citet{Ronneberger2015}, while the U-Net operates on small tiles only. To investigate the effects of different tile sizes, we train networks using square tiles with edge lengths 512~px and 256~px respectively. We apply data augmentation motivated by the techniques and parameters of \citet{Miao2021} and \citet{Bergs2020}. Data augmentation is popular especially in computer vision tasks, to get a robust model despite a limited amount of data. In order to investigate the effects of data augmentation in the application of wear area segmentation, we compare models trained on tiles without any augmentation, with moderate augmentation, and with strong augmentation.
Furthermore, the networks are trained using three different loss functions: The basic Cross Entropy (CE), the Focal Cross Entropy (FCE) addressing class imbalance \citep{Lin2017}, and a loss function based on Intersection over Union as a non-pixel-based loss function \citep{Beers2019}.
Worn areas of the cutting edge were partly covered by transfer material from the work piece. Thus, we labelled the wear area with regard to two types of wear: abrasive wear (referred to as A in the following), and adhesive wear in the form of transferred work piece material (referred to as M). We trained networks as a binary problem including both types in a joint wear class, and a multiclass problem with the two wear types as distinct classes.
In total, we investigate the performances of 72 different model configurations, depending on problem definition (binary or multiclass), tile size (edge length 512~px or 256~px), data augmentation (none, moderate, or full), loss function (CE, FCE, and IoU-based), and whether batch normalisation \citep{Ioffe2015} is used in the U-Net architecture.

For performance evaluation, the Intersection over Union and the Dice coefficient are popular choices.
We follow the recommendations by \citet{Mueller2022} to additionally state sensitivity (true positive rate; TPR) and specificity (true negative rate; TNR). For multiclass problems, micro- or macro-averaging scores over all classes may introduce biases towards the dominant classes (which is often the background), and has to be interpreted with caution. We are mainly interested in predicting the whole wear area correctly, rather than to distinguish the two wear types. Thus, we transform the multiclass predictions into binary predictions by joining the two wear types to one class. That way, the scores of the binary and the multiclass models are comparable. In addition, we compute the False negative background rate (FNBGR) with respect to the two wear types, stating which ratios of the two wear areas were falsely predicted to be no wear.

The paper is organised as follows. In Section \ref{sec:2}, we introduce and summarise the data set used for this study, discuss the data preparation and augmentation steps, formulate the problem statement, and define the loss functions. We close this section with a discussion of the implementation and training process. In Section \ref{sec:3}, we show the results obtained for the various combinations of loss functions and degrees of augmentation, while also comparing multiclass vs binary classifiers, while different tile sizes are considered. The best performing models are highlighted as well. Finally, in Section \ref{sec:4}, we conclude, and point out avenues for future research on this topic.   

\section{Materials and Methods}\label{sec:2}

\subsection{Data Description}

\begin{figure*}[ht, width=\textwidth]
\centering
\begin{subfigure}[t]{0.45\textwidth}
    \centering
    \frame{\includegraphics[width=\textwidth]{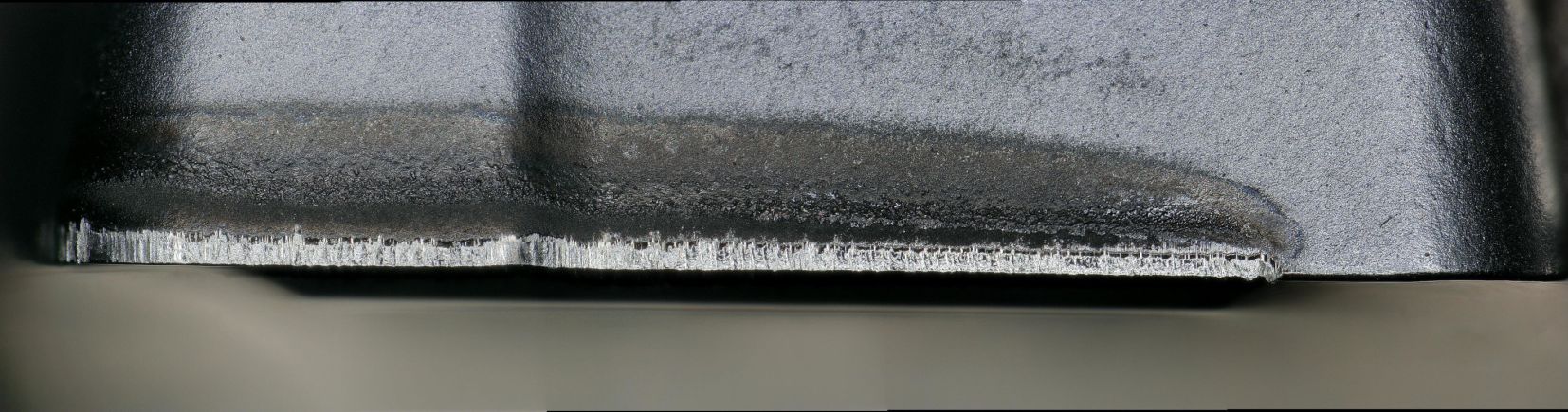}}
    \caption{} \label{fig:img}
\end{subfigure}
\hspace*{6mm}
\begin{subfigure}[t]{0.45\textwidth}
    \centering
    \frame{\includegraphics[width=\textwidth]{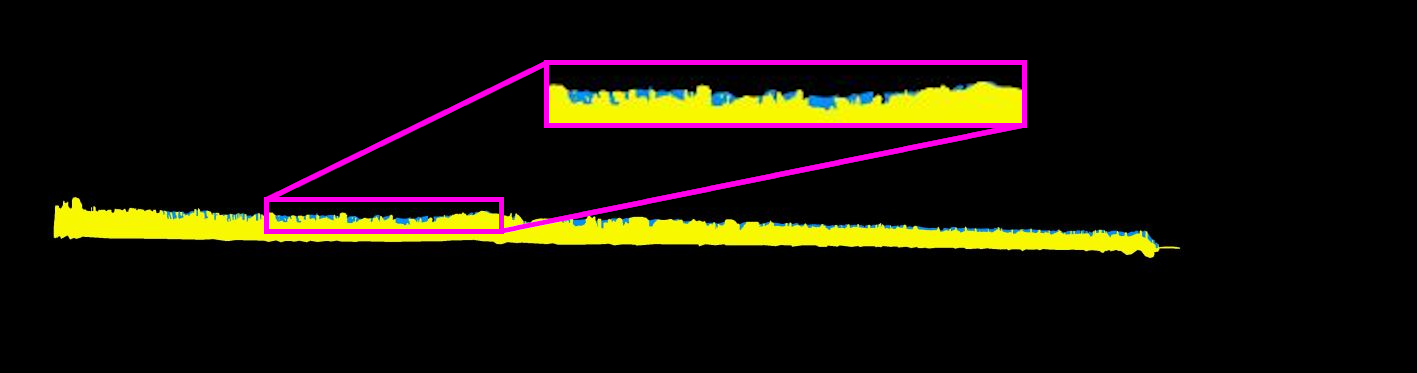}}
    \caption{} \label{fig:mask}
\end{subfigure}
\caption{(\subref{fig:img}) Light optical microscopy image of cutting insert, image done by digital microscope (VHX7000, Fa. Keyence); (\subref{fig:mask}) associated mask with abrasive wear area A in blue and area of transferred work piece material M in yellow; zoom for better visibility of the two wear types.} \label{fig:datasample}
\end{figure*}

The data comprise 24 microscopy images from the flank face of worn cutting inserts (XOMT 060304SN CTPP430, Fa. Ceratizit) clamped into a `MaxiDrill900' (Diameter 22, MD-900.3D.320.R.09-C32, Fa. Ceratizit), collected from several experiments with varying cutting velocity $v_c$ and feed per tool revolution $f_u$, but the same work piece material (X37CrMoV5-1). The images have a size of about $4700\times1200$~pixels per insert, where one pixel equals 0.781 to \SI{1.493}{\micro\metre}. In order to fully utilise the high quality images, we partitioned them into several smaller, square images, to fit the data into the Graphics Processing Unit's (GPU) memory. That way, we avoid information loss due to downsizing the images, and obtain a larger amount of training data. 
To each image belongs a manually labelled mask as ground truth, with two different labels for two distinct types of wear, abrasive wear A, and transferred work piece material M, as depicted in Fig.~\ref{fig:datasample}.

\subsection{Data Preparation}
A small set of 4 images from the original data set is put aside right at the beginning as a test set. This portion of images is used for evaluation of the trained models, hence no part of the image should be used in the model training process beforehand.

The remaining 20 images are used as training set. Images of the training set are cut into square tiles with a fixed edge length of $d$~pixels, with $d=512$ or $d=256$, to explore performance depending on tile size. Furthermore, for both tile sizes, the data is preprocessed in 3 ways: 1) tiles with the wear area centred vertically, 2) moderate augmentation of the tiles, 3) more intense augmentation of the tiles (called ``full augmentation'' in the following).
For the non-augmented data set, the images are partitioned into segments of width $d$. The segment is then cropped with respect to the vertical position of the wear area, such that the square tiles are centred vertically around the wear area. 
The augmentation methods include rotation, changing the image's contrast and brightness, adding a Gaussian blur, and shifting the window vertically/horizontally. Generally, shifting and rotating the image causes the augmented image to have some missing parts if the image size should remain the same. To avoid such empty parts within the image, we rotate and shift the image before cutting out a tile of the needed size from the larger, original image, whenever this is possible. This approach may cause overlaps between different tiles, wherefore the test set was chosen based on the original images beforehand to prevent information leakage between training and test set. Since additional augmentation methods are applied to the single tiles, also the overlapping parts appearing again on another tile are not identical but vary in contrast, brightness and/or blurring. Usually, a single image is augmented in several ways to produce more training data. Our approach, however, does not use one tile to generate several augmented samples, but is based on overlapping segments from the original image. 

\begin{figure*}[ht, width=\textwidth]
\centering
    \centering
     \includegraphics[clip, trim=0 0 0 0, width=.9\textwidth]{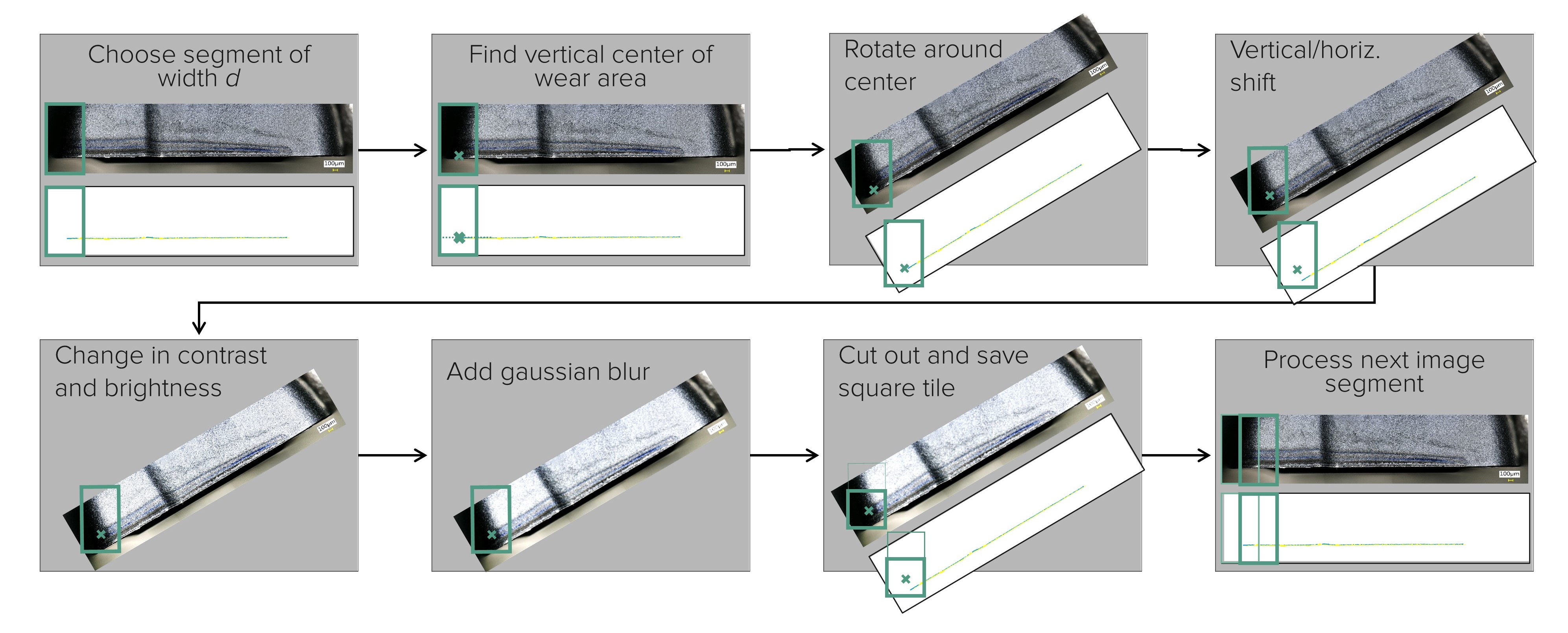}
\caption{Flowchart of data preprocessing with augmentation.}
\label{fig:flowaug}
\end{figure*}

The exact preparation workflow for the training sets with augmentation is depicted in Fig.~\ref{fig:flowaug}, with the parameters as stated in Table~\ref{tbl:dataaug}.

\begin{table*}[ht, width=\textwidth]
\centering
\caption{Data augmentation parameters. The parameters are sampled randomly from the given intervals.}
\label{tbl:dataaug}
    \begin{tabular}{l|ccc}
		\toprule
		                        & no augmentation & moderate augmentation & full augmentation \\
		\midrule
		rotation angle              & - & $[-30,30]$ & $[-90,90]$ \\
            horiz./vert. shift factor       & - & $[-0.15,0.15]$  & $[-0.3,0.3]$\\
		contrast scaling factor     & - &  $[0.9,1.1]$ & $[0.8,1.2]$ \\
            brightness scaling factor   & - &  $[0.9,1.1]$ & $[0.8,1.2]$ \\
		Gaussian blur radius        & - & - & $[0,1]$ \\
            stride to next segment      &  $d$  &  $d/2$  &  $d/2$  \\
	\end{tabular}

\end{table*}

The preprocessing results in six different training sets in total, i.e., non-augmented, moderately augmented and fully augmented, for two tile sizes respectively.
For edge length 512~px, this results in 156 tiles if no augmentation is applied, 308 tiles for moderate augmentation, and 306 tiles for full augmentation.
For edge length 256~px, this results in 300 tiles if no augmentation is applied, and 599 tiles for moderate and full augmentation respectively.

\subsection{Problem Definition \& Loss Functions}

The segmentation is performed for two different problem definitions: the first problem definition is formulated as a binary problem, treating the two different wear types as one class, which are all the pixels that are distinct from the non-wear pixels. The second problem definition is formulated as multiclass problem, with the background and the two wear types separately, resulting in three classes in total.
Depending on the problem definition, we use sigmoid (for binary) or softmax (for multiclass) as activation function in the output layer.
The sigmoid function 
$$S(x)=\frac{1}{1+\exp(-x)}$$
maps each scalar value $x$ to the interval $[0,1]$, which is then interpreted as the probability of a pixel belonging to the category of interest ($S(x)$ near 1) or not ($S(x)$ near 0).
The softmax function maps each component of the vector $\bm{x}$ of length $K$ to values between 0 and 1, summing up to 1:
$$\sigma(\bm{x})_j=\frac{\exp(x_j)}{\sum_{k=1}^{K}\exp(x_k)} \, .$$ 
With this, the component $j$ of $\sigma(\bm{x})$ can again be interpreted as the probability of a pixel belonging to class $j$.
The sigmoid or softmax function is applied to each value in the output layer, and thus yields a probability for every single pixel in the image.

The loss function needs to decide whether the predicted probability indicates membership to the correct class given by the ground truth mask. 
For the binary problem, the probability $\hat{p}=S(x)\in [0,1]$ is compared with the true value $y$, which equals 1 if the pixel belongs to the labelled area, and 0, if it does not belong to the labelled area. The predicted category for each pixel $\hat{y}$ is derived from the probability $\hat{p}$, with $\hat{p}\geq 0.5$ indicating $\hat{y}=1$, and $\hat{p}<0.5$ indicating $\hat{y}=0$.
For the multiclass problems, the probability $\hat{p}_j=\sigma(\bm{x})_j$ for each class $j=1,...,K$ is compared with the true $y_j$, which equals 1, if the pixel belongs to the class $j$, and 0, if it belongs to a class $k\neq j$. The prediction $\hat{\bm{y}}$ for a pixel is given as $\hat{y}_j=1$ for $j=argmax_{k=1,...,K} \sigma(\bm{x})_k$, and $\hat{y}_k=0$ for $k\neq j$.

In total, three different loss functions are explored for both, the binary and the multiclass problem: 

\subsubsection*{Cross Entropy}
The standard loss function for image segmentation tasks is the Cross Entropy (CE). The categorical Cross Entropy is given as 
$$
\text{CE} = - \sum_{k=1}^K y_k \cdot \log(\hat{p}_k) \, .
$$
For two classes, this can be reformulated using $y_1:=y$, $\hat{p}_1:=\hat{p}$, $y_2:=1-y$ and $\hat{p}_2:=1-\hat{p}$, resulting in the binary Cross Entropy:
$$
\text{CE}_B = - y \cdot \log(\hat{p}) - (1-y) \cdot \log(1-\hat{p})\, .
$$

In this paper, we use the implementations of the (categorical) cross entropy loss as described in \citet{keras1, keras2}, provided by TensorFlow, \citet{tensorflow} and Keras, \citet{keras}. 
    
\subsubsection*{Focal Cross Entropy}
The Focal Cross Entropy (FCE), as proposed by \citet{Lin2017}, reduces the impact of well-classified samples on the total loss, focusing on samples that are hard to classify. This is realised via additional factors in the Cross Entropy loss---the summand is rescaled by a factor of $(1-\hat{p}_k)^{\gamma}$, with the focusing parameter $\gamma\geq0$:
$$
\text{FCE} = - \sum_{k=1}^K y_k \cdot (1-\hat{p}_k)^{\gamma} \cdot \log(\hat{p}_k) \, .
$$
The binary case can be reformulated analogously, as for the standard Cross Entropy, we have:
$$
\text{FCE}_B = - y \cdot (1-\hat{p})^{\gamma} \cdot \log(\hat{p}) - (1-y) \cdot \hat{p}^{\gamma} \cdot \log(1-\hat{p})\, .
$$
For $\gamma=0$, this reduces to the standard Cross Entropy. A larger value of $\gamma$ leads to a faster decay of the factor in the interval $[0,1]$, and thus a more down-weighting of easy-to-classify samples. In this work, the default value $\gamma=2$ is used.

In this paper, we use the implementations of the binary focal cross entropy loss as described in \citet{keras3}, and the sparse categorical focal loss as described in \citet{keras4} for the multiclass problem, with both functions provided by TensorFlow, \citet{tensorflow} and Keras, \citet{keras}.

\subsubsection*{IoU-based Loss}
The Intersection over Union is defined as the intersection of true and predicted class over the union of true and predicted class. For a binary classification problem, this can be formulated as
$$
\text{IoU} = \frac{\text{TP}}{\text{FP}+\text{FN}+\text{TP}} \, ,
$$
with TP indicating the sum of pixels that are correctly predicted to belonging to the class, FP incorrectly predicted to belonging to the class, and FN incorrectly predicted to not belonging to the class of interest.
To get a differentiable function usable in gradient descent, the probabilities $\hat{p}$ are used instead of the predicted categories $\hat{y}$, as stated in \citet{Miao2021}, i.e., $\text{TP}=\sum y \cdot \hat{y}$, $\text{FP}=\sum (1-y) \cdot \hat{y}$, and $\text{FN}=\sum y \cdot (1-\hat{y})$. 
With this, the IoU-based loss for the binary problem is defined as
$$
L_{\text{IoU},B} = 1 - \text{IoU} \, .
$$

For the multiclass problem, the IoU-based loss computed via a weighted average of the IoU of each category $k$, with weights $w_k$:
$$
L_{\text{IoU}} = 1 - \frac{1}{K} \sum_{k=1}^K w_k \text{IoU}_k \, .
$$

This work uses the weight 0.2 for the background class, and weight 1.4 for the two wear classes respectively. Since not all classes have to be present in a tile, the denominator can become zero, resulting in a nan value for the IoU of that specific class. This nan values are set to 1, because they reflect 100\% true negative, and thus a correct classification.

\subsection{Implementation \& Training}

The implementation is realised with TensorFlow version 2.11.0, and the code associated to this paper can be found on the GitHub repository \citet{Schlager2022}, which uses the tiler \citet{Tiler2022}. 
The U-Net used in this work is built upon the basic U-Net architecture as proposed by \citep{Ronneberger2015}: Four encoder blocks are followed by a convolutional block, and then four decoder blocks. The convolutional layers in the network have kernel size (3,3) and use ``same'' padding; the rest of the parameters use TensorFlow's default settings. We optionally use batch normalisation right after the convolutional layers, before ReLu activation, and compare the model's performances with and without batch normalisation. 

5-fold cross validation with random re-initialisation of the weights is used in order to explore robustness in training with respect to changing initialisation and training-validation-splits. The training is performed using the Adam optimiser with an initial learning rate of $1e-4$, which is reduced using \textit{tf.keras.callbacks.ReduceLROnPlateau} with a factor of 0.9, a patience of 5, $\text{min\_delta}=0.005$, till the learning rate falls to a minimal rate of $1e-6$. The rest of the hyperparameters are set to default, \ie, $\text{beta\_1}=0.9$, $\text{beta\_2}=0.999$, and $\text{epsilon}=1e-07$.
Training is executed for at most 300 epochs, with early stopping if validation loss did not improve over 20 epochs, becoming active after the first 60 epochs; the model having the lowest validation loss till stopping is then restored and saved. However, if at the end of training the validation loss is not below a threshold of 0.4, the training is assumed to have failed, and the model is reinitialised and trained again for that fold.
The training is performed in parallel on two GPUs (Tesla P100-16 GB), allowing a batch size of 16 for $d=512$, and a batch size of 32 for $d=256$.

For each of the six training sets, the U-Net is trained with and without batch normalisation, as binary, and as multiclass problem, using the three different loss functions respectively. Thus, training with 5-fold cross validation is performed for 72 different combinations of settings. 

\section{Results and Discussion}\label{sec:3}

For evaluation, the models are applied to non-augmented tiles coming from the four unseen images in the test set. However, the trained models can only be applied to tiles having the same size as they were trained on. For edge length $d=512$~px, the test set comprises 32 tiles, for $d=256$~px it comprises 59 tiles.
Analogously, as for the training set, only tiles exhibiting some wear are used. Therefore, the models using smaller tile size have in total a smaller proportion of background during training and evaluation. As a consequence, the evaluation scores are based on different class proportions for the two tile sizes, and are thus not directly comparable.
The IoU, Dice coefficient, sensitivity (TPR), and specificity (TNR) are calculated for each tile in the test set, and then averaged to get one score per model fold. 
While for the binary problem definition the scores refer to the joined wear class only by design, scores for multiclass problems generally also include the background class. To be able to compare the performance of the binary with the multiclass problem, we transform the predictions of the multiclass model to binary predictions for evaluation. That way, the U-Net was trained on the wear types separately, but evaluated as one class, analogous to the binary problem definition.

\begin{figure*}[!ht, width=\textwidth]
    \begin{subfigure}[t]{\textwidth}
    \centering
        \subcaptionbox{Binary model of tiles with edge length 512 px}{\includegraphics[clip, trim=0 10 0 35, width=.49\textwidth]{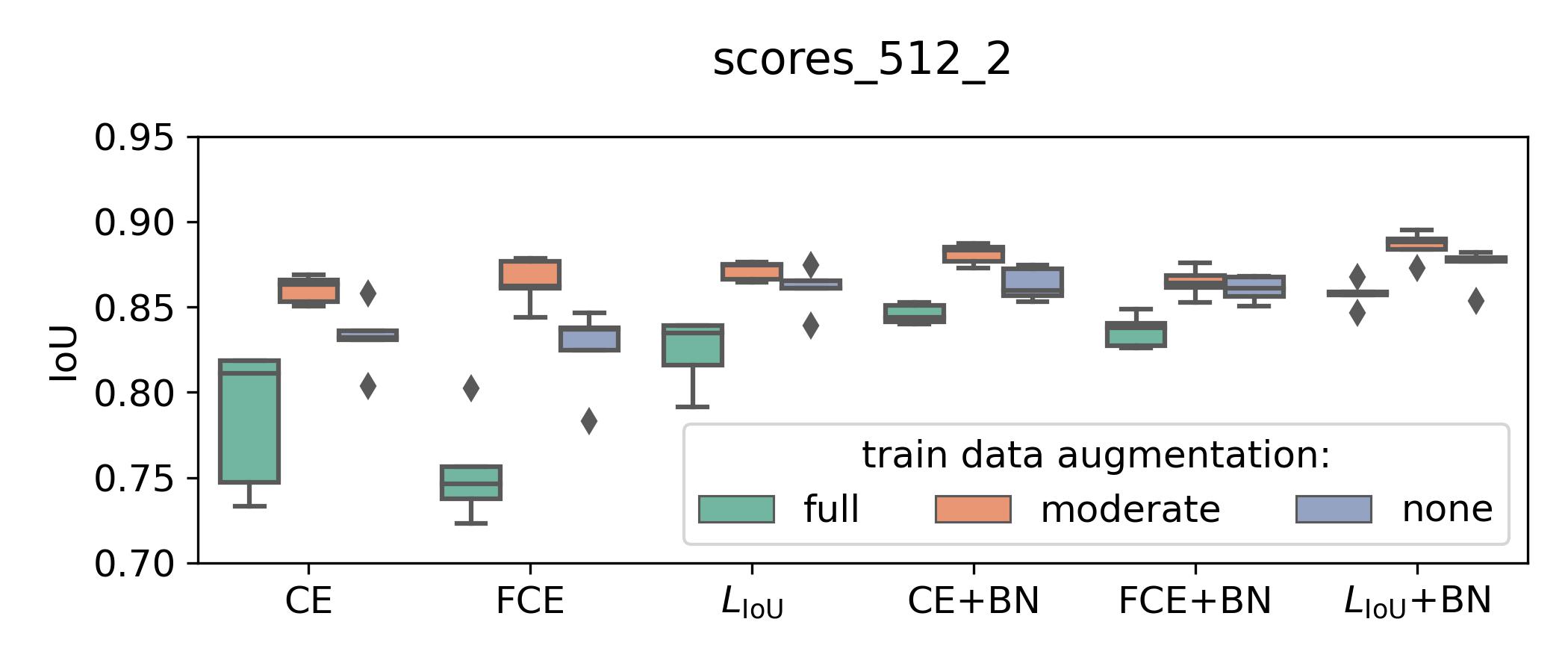}}
        \hfill
        \subcaptionbox{multiclass model of tiles with edge length 512 px}{\includegraphics[clip, trim=0 10 0 35, width=.49\textwidth]{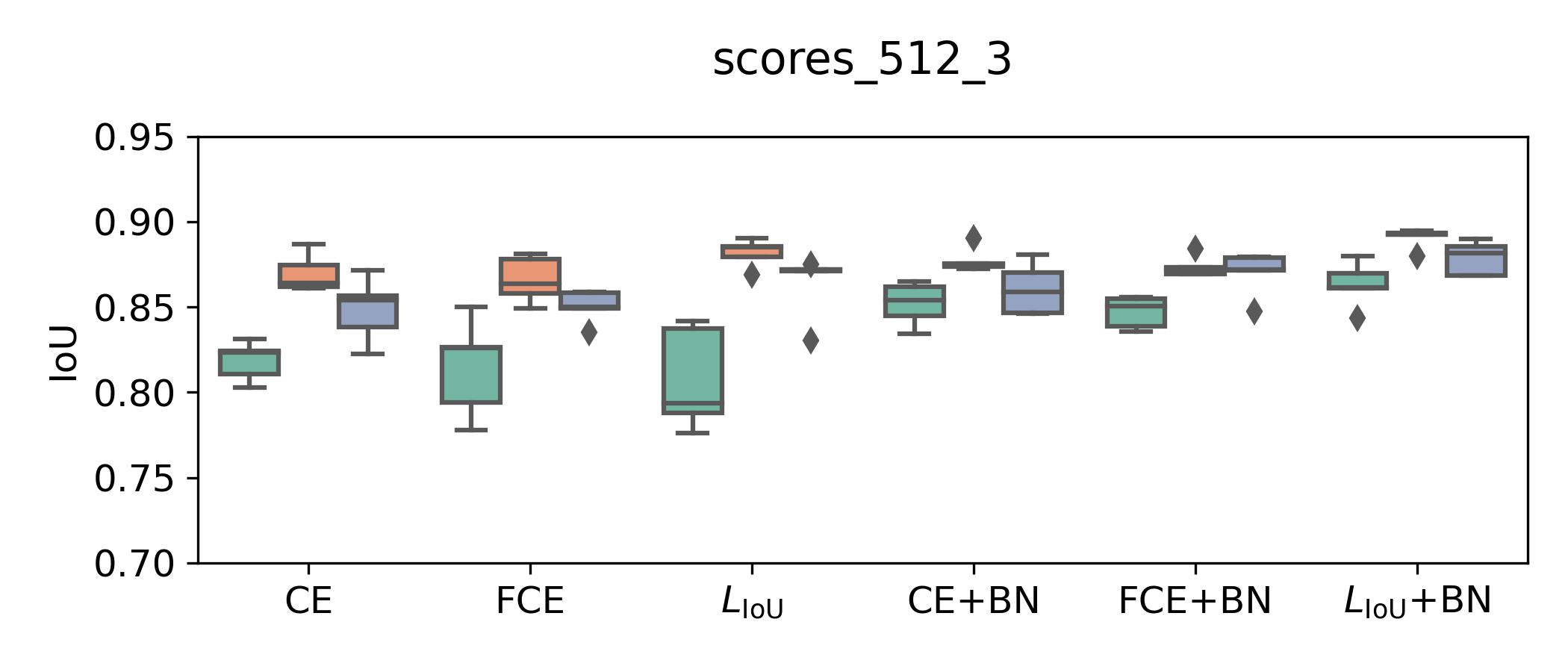}}
    \vspace*{4mm}
    \end{subfigure}
    \begin{subfigure}[t]{\textwidth}
    \centering
        \subcaptionbox{Binary model of tiles with edge length 256 px}{\includegraphics[clip, trim=0 10 0 35, width=.49\textwidth]{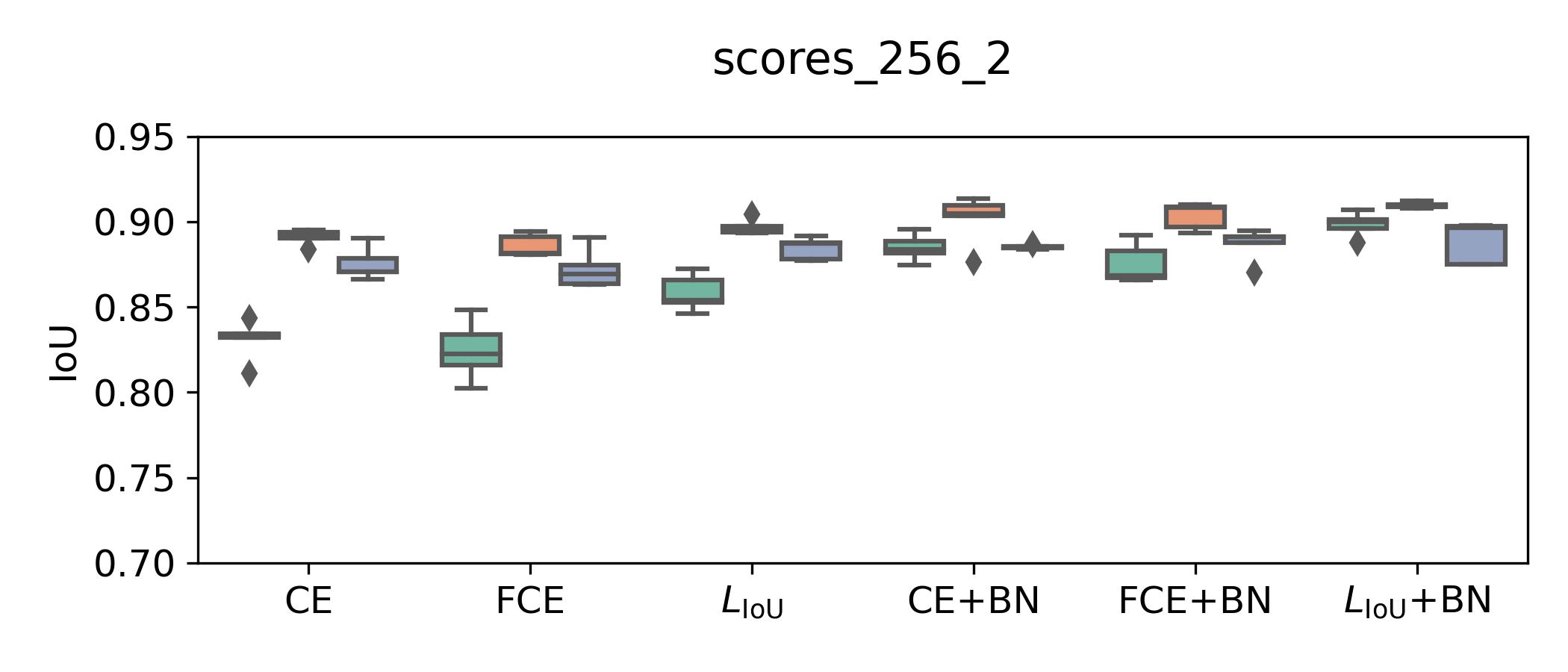}}
        \hfill
        \subcaptionbox{multiclass model of tiles with edge length 256 px}{\includegraphics[clip, trim=0 10 0 35, width=.49\textwidth]{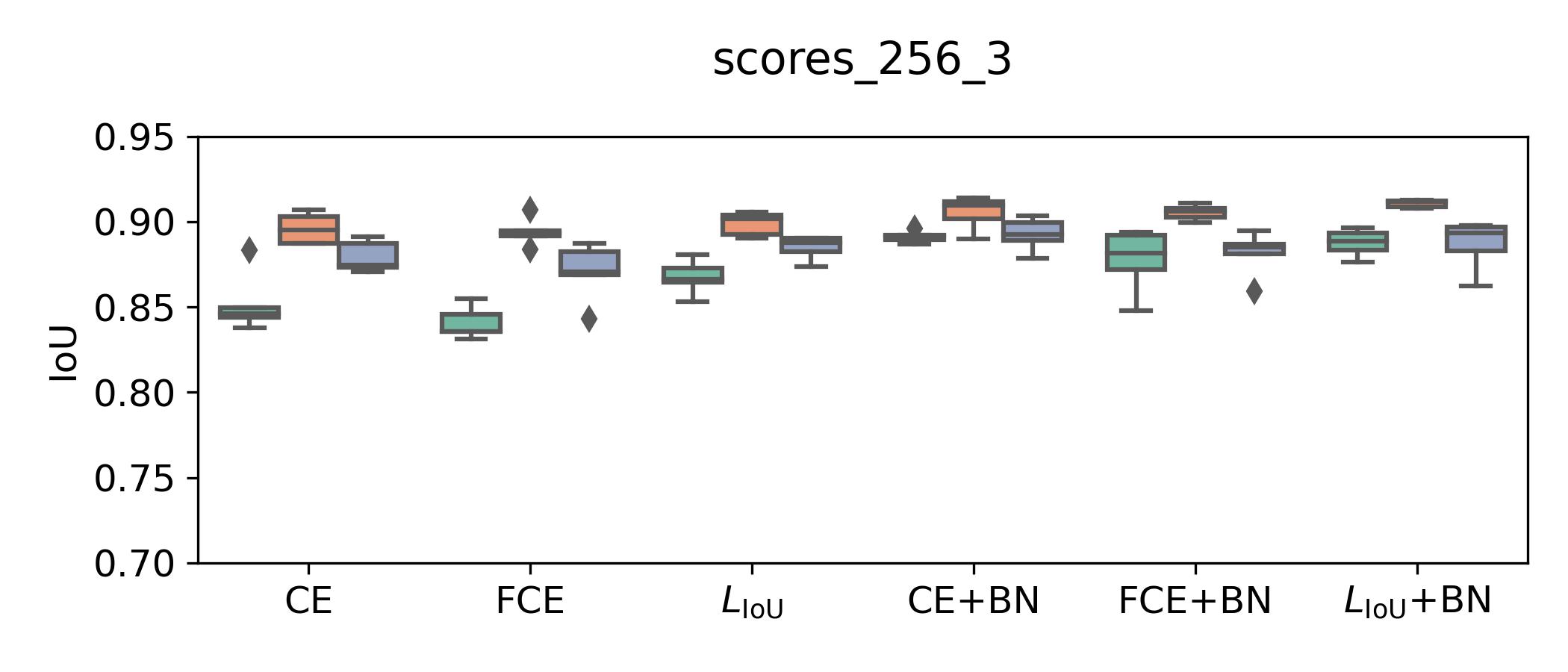}}
    \vspace*{2mm}
    \end{subfigure}
\caption{Distribution of IoU of 5-fold cross validation of the models trained with loss functions CE, FCE and $L_{\text{IoU}}$, with and without batch normalisation (BN), on data with full, moderate, and no augmentation. The test data are unseen and not augmented tiles with edge length 512~px and 256~px respectively. }
\label{fig:Boxplots_scores}
\end{figure*}

Fig.~\ref{fig:Boxplots_scores} depicts the performance of the models as boxplots of the average IoUs of the five folds.
Each plot shows the test scores for the models trained with the data sets of different augmentation levels (denoted by different colours), for the three different loss functions and with/without batch normalisation (denoted by the x-axis). 
Varying the training set while fixing all the other settings, \ie, tile size, loss function, use of batch normalisation, and whether binary or multiclass classification is used, we can observe the IoU being lowest when training was performed with fully augmented data, and highest when training was performed with data of moderate augmentation.
Another striking aspect observable in the figures is the higher level of the IoU for the multiclass models compared to the associated binary models. Furthermore, the median IoU tends to be higher for the models with batch normalisation, compared to the associated models without batch normalisation.

A more detailed evaluation is presented in Tables~\ref{tbl:scores512} and~\ref{tbl:scores256}, which list the median and interquartile range (IQR) of IoU, Dice coefficient, TPR, TNR, and FNBGR of both wear types of the five runs from cross validation of the respective models. The scores are computed based on the binary labels, \ie, wear or not wear, disregarding the distinction of the multiclass prediction. 
The False negative background rate (FNBGR) of the predictions with respect to the two wear types M (transferred work piece material), and A (abrasive), state the ratio of the pixels of wear M, or A, which are falsely predicted to be of no wear, but neglecting whether the multiclass model predicts wear M or wear A for the pixels. 

For each metric, the best performing score for binary and multiclass models is highlighted in a bold face font. For tiles of edge length 512~px, for binary and multiclass problems, the models with batch normalisation and trained with $L_{\text{IoU}}$ loss function on moderately augmented training data shows best IoU, Dice, and TPR. The TNR is on a high level for all models, indicating that most of the background can be easily identified as such. For the multiclass models, FNBGR~M and FNBGR~A are lowest again for the model with batch normalisation trained with $L_{\text{IoU}}$ on moderately augmented training data. The area of wear A is in general much smaller than the area of wear M, and a lot more of area M, namely 11.7\% compared to 4.2\%, is falsely predicted as no wear area. The FNBGR for the two wear types is higher for the binary models in comparison to the multiclass models, indicating that the distinction of the wear types may help training.
For tiles of edge length 256~px, Table~\ref{tbl:scores256} shows very similar results: IoU, Dice, TPR, and FNBGR~M for binary and multiclass models are best with batch normalisation, trained with IoU-based loss on training data with moderate augmentation. While the best median FNBGR~M for binary and multiclass model are both 3.5\%, the best FNBGR~A is lower for the multiclass model (9.6\%) than for the binary model (10.7\%).

\begin{table*}[!htb, width=\textwidth]
\centering
\caption{Median and interquartile range of test scores for models using tiles with edge length 512 px.} \label{tbl:scores512}
\resizebox{\textwidth}{!}{\begin{tabular}{lll|cccccc}
\toprule
                 &     &      &            IoU &           Dice &            TPR &            TNR &        FNBGR M &        FNBGR A \\
mode & loss & train data aug. &                &                &                &                &                &                \\
\toprule
\multirow{9}{*}{binary} & \multirow{3}{*}{CE} & full &  0.811 (0.071) &  0.875 (0.050) &  0.894 (0.023) &  0.992 (0.003) &  0.093 (0.037) &  0.219 (0.070) \\
                 &     & moderate &  0.863 (0.013) &  0.920 (0.016) &  0.919 (0.016) &  0.995 (0.001) &  0.068 (0.021) &  0.175 (0.030) \\
                 &     & none &  0.832 (0.005) &  0.893 (0.011) &  0.919 (0.017) &  0.992 (0.004) &  0.068 (0.012) &  0.160 (0.043) \\
\cline{2-9}
                 & \multirow{3}{*}{FCE} & full &  0.747 (0.018) &  0.828 (0.015) &  0.888 (0.040) &  0.990 (0.003) &  0.090 (0.043) &  0.197 (0.098) \\
                 &     & moderate &  0.862 (0.016) &  0.920 (0.011) &  0.930 (0.022) &  0.995 (0.000) &  0.063 (0.010) &  0.142 (0.068) \\
                 &     & none &  0.837 (0.013) &  0.902 (0.008) &  0.922 (0.040) &  0.993 (0.001) &  0.063 (0.030) &  0.144 (0.136) \\
\cline{2-9}
                 & \multirow{3}{*}{$L_{\text{IoU}}$} & full &  0.835 (0.023) &  0.892 (0.011) &  0.934 (0.038) &  0.994 (0.002) &  0.053 (0.031) &  0.172 (0.058) \\
                 &     & moderate &  0.875 (0.009) &  0.925 (0.006) &  0.935 (0.021) &  0.995 (0.000) &  0.058 (0.020) &  \bfseries{0.139 (0.022)} \\
                 &     & none &  0.861 (0.005) &  0.917 (0.003) &  0.926 (0.005) &  0.995 (0.001) &  0.060 (0.007) &  0.153 (0.020) \\
\cline{1-9}
\cline{2-9}
\multirow{9}{*}{binary + BN} & \multirow{3}{*}{CE} & full &  0.844 (0.010) &  0.902 (0.012) &  0.899 (0.017) &  0.995 (0.001) &  0.086 (0.021) &  0.235 (0.011) \\
                 &     & moderate &  0.883 (0.008) &  0.930 (0.015) &  0.936 (0.010) &  0.995 (0.000) &  0.054 (0.010) &  0.143 (0.004) \\
                 &     & none &  0.860 (0.016) &  0.911 (0.013) &  0.921 (0.020) &  \bfseries{0.996 (0.000)} &  0.057 (0.026) &  0.207 (0.019) \\
\cline{2-9}
                 & \multirow{3}{*}{FCE} & full &  0.838 (0.013) &  0.895 (0.010) &  0.896 (0.026) &  0.995 (0.000) &  0.091 (0.024) &  0.222 (0.043) \\
                 &     & moderate &  0.864 (0.007) &  0.911 (0.008) &  0.928 (0.020) &  0.995 (0.001) &  0.067 (0.020) &  0.154 (0.037) \\
                 &     & none &  0.861 (0.011) &  0.917 (0.008) &  0.913 (0.012) &  \bfseries{0.996 (0.001)} &  0.060 (0.012) &  0.231 (0.051) \\
\cline{2-9}
                 & \multirow{3}{*}{$L_{\text{IoU}}$} & full &  0.858 (0.002) &  0.916 (0.002) &  0.933 (0.008) &  0.994 (0.000) &  0.056 (0.001) &  0.148 (0.043) \\
                 &     & moderate &  \bfseries{0.888 (0.006)} &  \bfseries{0.935 (0.005)} &  \bfseries{0.943 (0.010)} &  0.995 (0.001) &  0.051 (0.011) &  0.142 (0.015) \\
                 &     & none &  0.878 (0.002) &  0.929 (0.002) &  0.937 (0.002) &  0.995 (0.000) &  \bfseries{0.047 (0.003)} &  0.170 (0.028) \\
\cline{1-9}
\cline{2-9}
\toprule
\multirow{9}{*}{multiclass} & \multirow{3}{*}{CE} & full &  0.823 (0.014) &  0.887 (0.009) &  0.903 (0.059) &  0.992 (0.004) &  0.068 (0.024) &  0.231 (0.131) \\
                 &     & moderate &  0.864 (0.013) &  0.919 (0.008) &  0.935 (0.007) &  0.995 (0.001) &  0.054 (0.005) &  0.144 (0.027) \\
                 &     & none &  0.854 (0.018) &  0.909 (0.021) &  0.899 (0.022) &  0.995 (0.001) &  0.082 (0.027) &  0.202 (0.042) \\
\cline{2-9}
                 & \multirow{3}{*}{FCE} & full &  0.826 (0.032) &  0.888 (0.032) &  0.932 (0.014) &  0.993 (0.004) &  0.056 (0.010) &  0.167 (0.008) \\
                 &     & moderate &  0.864 (0.020) &  0.920 (0.015) &  0.925 (0.026) &  \bfseries{0.996 (0.000)} &  0.063 (0.032) &  0.219 (0.078) \\
                 &     & none &  0.850 (0.009) &  0.912 (0.005) &  0.921 (0.017) &  0.994 (0.001) &  0.060 (0.008) &  0.217 (0.038) \\
\cline{2-9}
                 & \multirow{3}{*}{$L_{\text{IoU}}$} & full &  0.794 (0.049) &  0.862 (0.040) &  0.916 (0.052) &  0.993 (0.002) &  0.065 (0.047) &  0.224 (0.064) \\
                 &     & moderate &  0.885 (0.006) &  0.933 (0.005) &  0.939 (0.016) &  \bfseries{0.996 (0.000)} &  0.051 (0.011) &  0.143 (0.042) \\
                 &     & none &  0.872 (0.001) &  0.925 (0.002) &  0.928 (0.012) &  0.995 (0.001) &  0.063 (0.007) &  0.132 (0.021) \\
\cline{1-9}
\cline{2-9}
\multirow{9}{*}{multiclass + BN} & \multirow{3}{*}{CE} & full &  0.854 (0.017) &  0.906 (0.012) &  0.910 (0.023) &  0.995 (0.001) &  0.073 (0.024) &  0.176 (0.014) \\
                 &     & moderate &  0.875 (0.002) &  0.918 (0.004) &  0.923 (0.009) &  \bfseries{0.996 (0.000)} &  0.068 (0.003) &  0.145 (0.012) \\
                 &     & none &  0.859 (0.023) &  0.916 (0.023) &  0.894 (0.023) &  \bfseries{0.996 (0.001)} &  0.071 (0.024) &  0.223 (0.039) \\
\cline{2-9}
                 & \multirow{3}{*}{FCE} & full &  0.851 (0.016) &  0.902 (0.015) &  0.908 (0.018) &  0.995 (0.001) &  0.084 (0.023) &  0.224 (0.074) \\
                 &     & moderate &  0.872 (0.004) &  0.918 (0.000) &  0.924 (0.006) &  0.995 (0.001) &  0.073 (0.004) &  0.147 (0.017) \\
                 &     & none &  0.872 (0.007) &  0.926 (0.004) &  0.928 (0.025) &  \bfseries{0.996 (0.001)} &  0.055 (0.022) &  0.227 (0.074) \\
\cline{2-9}
                 & \multirow{3}{*}{$L_{\text{IoU}}$} & full &  0.861 (0.009) &  0.916 (0.009) &  0.930 (0.015) &  0.994 (0.001) &  0.059 (0.020) &  0.136 (0.019) \\
                 &     & moderate &  \bfseries{0.892 (0.001)} &  \bfseries{0.938 (0.002)} &  \bfseries{0.951 (0.002)} &  0.995 (0.000) &  \bfseries{0.042 (0.003)} &  \bfseries{0.117 (0.011)} \\
                 &     & none &  0.881 (0.017) &  0.932 (0.012) &  0.937 (0.032) &  0.995 (0.001) &  0.048 (0.023) &  0.180 (0.070) \\
\bottomrule
\end{tabular}}
\end{table*}

\begin{table*}[!htb, width=\textwidth]
\centering
\caption{Median and interquartile range of test scores for models using tiles with edge length 256 px.} \label{tbl:scores256}
\resizebox{\textwidth}{!}{\begin{tabular}{lll|cccccc}
\toprule
                 &     &      &            IoU &           Dice &            TPR &            TNR &        FNBGR M &        FNBGR A \\
mode & loss & train data aug. &                &                &                &                &                &                \\
\toprule
\multirow{9}{*}{binary} & \multirow{3}{*}{CE} & full &  0.834 (0.002) &  0.891 (0.012) &  0.898 (0.018) &  0.987 (0.002) &  0.087 (0.012) &  0.183 (0.061) \\
                 &     & moderate &  0.893 (0.003) &  0.938 (0.002) &  0.938 (0.006) &  0.988 (0.001) &  0.052 (0.010) &  0.137 (0.012) \\
                 &     & none &  0.871 (0.008) &  0.926 (0.007) &  0.941 (0.015) &  0.985 (0.002) &  0.048 (0.017) &  0.185 (0.031) \\
\cline{2-9}
                 & \multirow{3}{*}{FCE} & full &  0.823 (0.018) &  0.885 (0.010) &  0.901 (0.021) &  0.983 (0.004) &  0.084 (0.018) &  0.120 (0.061) \\
                 &     & moderate &  0.882 (0.010) &  0.933 (0.007) &  0.942 (0.005) &  0.986 (0.002) &  0.046 (0.006) &  0.141 (0.029) \\
                 &     & none &  0.869 (0.011) &  0.925 (0.008) &  0.936 (0.005) &  0.983 (0.001) &  0.053 (0.007) &  0.151 (0.028) \\
\cline{2-9}
                 & \multirow{3}{*}{$L_{\text{IoU}}$} & full &  0.854 (0.013) &  0.909 (0.014) &  0.929 (0.004) &  0.984 (0.001) &  0.063 (0.007) &  \bfseries{0.107 (0.037)} \\
                 &     & moderate &  0.896 (0.004) &  0.942 (0.003) &  \bfseries{0.953 (0.007)} &  0.986 (0.002) &  0.039 (0.004) &  0.119 (0.016) \\
                 &     & none &  0.887 (0.010) &  0.935 (0.008) &  0.939 (0.012) &  0.986 (0.001) &  0.052 (0.011) &  0.126 (0.009) \\
\cline{1-9}
\cline{2-9}
\multirow{9}{*}{binary + BN} & \multirow{3}{*}{CE} & full &  0.884 (0.007) &  0.931 (0.009) &  0.947 (0.007) &  0.986 (0.003) &  0.038 (0.014) &  0.154 (0.051) \\
                 &     & moderate &  0.905 (0.006) &  0.946 (0.006) &  0.946 (0.013) &  \bfseries{0.990 (0.001)} &  0.045 (0.013) &  0.124 (0.011) \\
                 &     & none &  0.885 (0.001) &  0.933 (0.002) &  0.930 (0.027) &  0.988 (0.003) &  0.046 (0.012) &  0.221 (0.049) \\
\cline{2-9}
                 & \multirow{3}{*}{FCE} & full &  0.869 (0.016) &  0.926 (0.009) &  0.911 (0.020) &  \bfseries{0.990 (0.001)} &  0.051 (0.007) &  0.279 (0.084) \\
                 &     & moderate &  0.908 (0.012) &  0.950 (0.013) &  0.942 (0.010) &  \bfseries{0.990 (0.001)} &  0.049 (0.012) &  0.150 (0.014) \\
                 &     & none &  0.891 (0.004) &  0.938 (0.003) &  0.941 (0.004) &  0.987 (0.001) &  0.045 (0.005) &  0.172 (0.012) \\
\cline{2-9}
                 & \multirow{3}{*}{$L_{\text{IoU}}$} & full &  0.900 (0.005) &  0.945 (0.002) &  0.951 (0.006) &  0.987 (0.002) &  0.037 (0.002) &  0.141 (0.010) \\
                 &     & moderate &  \bfseries{0.909 (0.001)} &  \bfseries{0.951 (0.001)} &  \bfseries{0.953 (0.006)} &  0.988 (0.001) &  \bfseries{0.035 (0.003)} &  0.131 (0.026) \\
                 &     & none &  0.896 (0.022) &  0.941 (0.019) &  0.947 (0.014) &  0.985 (0.005) &  0.039 (0.009) &  0.147 (0.032) \\
\cline{1-9}
\cline{2-9}
\toprule
\multirow{9}{*}{multiclass} & \multirow{3}{*}{CE} & full &  0.846 (0.006) &  0.900 (0.011) &  0.917 (0.033) &  0.988 (0.005) &  0.077 (0.036) &  0.173 (0.075) \\
                 &     & moderate &  0.895 (0.016) &  0.942 (0.011) &  0.942 (0.013) &  0.988 (0.001) &  0.042 (0.001) &  0.148 (0.059) \\
                 &     & none &  0.874 (0.014) &  0.923 (0.014) &  0.934 (0.010) &  0.988 (0.001) &  0.052 (0.006) &  0.187 (0.050) \\
\cline{2-9}
                 & \multirow{3}{*}{FCE} & full &  0.836 (0.010) &  0.896 (0.009) &  0.911 (0.014) &  0.984 (0.005) &  0.070 (0.015) &  0.153 (0.024) \\
                 &     & moderate &  0.893 (0.003) &  0.939 (0.001) &  0.933 (0.011) &  \bfseries{0.990 (0.001)} &  0.048 (0.004) &  0.150 (0.032) \\
                 &     & none &  0.870 (0.013) &  0.922 (0.012) &  0.932 (0.025) &  0.989 (0.001) &  0.044 (0.015) &  0.207 (0.047) \\
\cline{2-9}
                 & \multirow{3}{*}{$L_{\text{IoU}}$} & full &  0.866 (0.008) &  0.920 (0.005) &  0.920 (0.013) &  0.986 (0.001) &  0.070 (0.016) &  0.113 (0.028) \\
                 &     & moderate &  0.902 (0.011) &  0.946 (0.007) &  0.948 (0.015) &  0.988 (0.001) &  0.038 (0.004) &  0.147 (0.059) \\
                 &     & none &  0.888 (0.008) &  0.937 (0.006) &  0.936 (0.017) &  0.986 (0.005) &  0.058 (0.006) &  \bfseries{0.096 (0.036)} \\
\cline{1-9}
\cline{2-9}
\multirow{9}{*}{multiclass + BN} & \multirow{3}{*}{CE} & full &  0.891 (0.003) &  0.938 (0.001) &  0.933 (0.005) &  0.989 (0.001) &  0.046 (0.001) &  0.184 (0.066) \\
                 &     & moderate &  0.909 (0.010) &  0.950 (0.007) &  0.949 (0.010) &  0.989 (0.001) &  0.041 (0.013) &  0.143 (0.005) \\
                 &     & none &  0.892 (0.011) &  0.937 (0.008) &  0.930 (0.010) &  \bfseries{0.990 (0.001)} &  0.048 (0.014) &  0.213 (0.054) \\
\cline{2-9}
                 & \multirow{3}{*}{FCE} & full &  0.882 (0.020) &  0.933 (0.012) &  0.920 (0.023) &  \textbf{0.990 (0.001)} &  0.055 (0.009) &  0.255 (0.068) \\
                 &     & moderate &  0.906 (0.005) &  0.947 (0.005) &  0.944 (0.009) &  \bfseries{0.990 (0.002)} &  0.046 (0.002) &  0.132 (0.069) \\
                 &     & none &  0.885 (0.006) &  0.935 (0.003) &  0.922 (0.016) &  0.988 (0.003) &  0.047 (0.016) &  0.223 (0.053) \\
\cline{2-9}
                 & \multirow{3}{*}{$L_{\text{IoU}}$} & full &  0.888 (0.010) &  0.937 (0.007) &  0.945 (0.003) &  0.989 (0.002) &  0.040 (0.006) &  0.140 (0.028) \\
                 &     & moderate &  \bfseries{0.912 (0.003)} &  \bfseries{0.952 (0.002)} &  \bfseries{0.955 (0.009)} &  0.988 (0.001) &  \bfseries{0.035 (0.005)} &  0.122 (0.030) \\
                 &     & none &  0.893 (0.014) &  0.940 (0.009) &  0.948 (0.007) &  0.988 (0.003) &  0.038 (0.009) &  0.153 (0.007) \\
\bottomrule
\end{tabular}}
\end{table*}

Based on the evaluation of the cross validation, and for each tile size, we pick the best performing binary and multiclass settings with respect to IoU and Dice, \ie, U-Net with batch normalisation layers, and $L_{\text{IoU}}$ as loss function, trained on data with moderate augmentation. 
The models are trained on 90\% of the training data, while the other 10\% are used for early stopping. Performance scores on the respective test sets are shown in Table~\ref{tbl:scoresfinal}. For both tile sizes, IoU, Dice TPR and TNR are very similar. However, while for smaller tiles the FNBGRs are smaller for the multiclass models, for tiles of edge length 512, FNBGR~M is larger for the multiclass model than for the binary model.

\begin{table*}[!ht, width=\textwidth]
\centering
\caption{Test scores of final models for tiles with edge length 512 px and 256 px respectively.} \label{tbl:scoresfinal}
\begin{tabular}{lc|cccccc}
\toprule
        &  mode  &    IoU   &   Dice     &    TPR    &    TNR   &   FNBGR M    &   FNBGR A   \\
\midrule
\multirow{2}{*}{512}
   &  binary        &   0.886  &   0.934    &   0.950   &   0.995  &  0.042  &  0.129 \\
   &  multiclass   &   0.884  &   0.930    &   0.944   &   0.994  &  0.051  &  0.107 \\
   
\midrule
\multirow{2}{*}{256}
   &  binary        &   0.904  &   0.947    &   0.948   &   0.988  &  0.037  &  0.140 \\
   &  multiclass   &   0.900  &   0.945    &   0.957   &   0.985  &  0.030  &  0.133 \\
\bottomrule
\end{tabular}
\end{table*}

\begin{figure*}[!ht, width=\textwidth]
    \begin{subfigure}[t]{\textwidth}
    \centering
        \includegraphics[width=.32\textwidth]{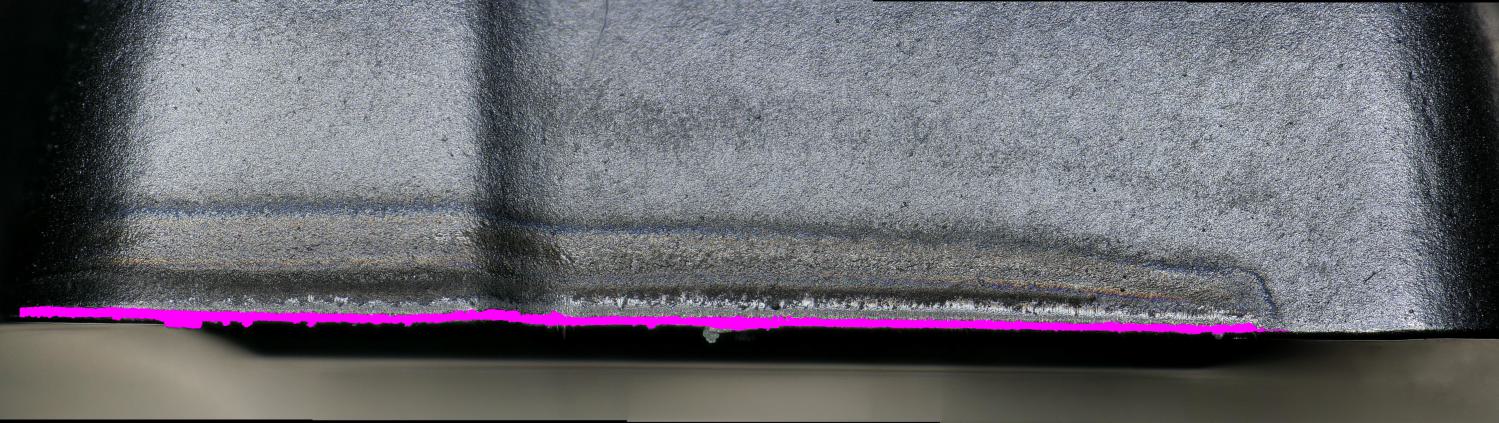}
        \hfill
        \includegraphics[width=.32\textwidth]{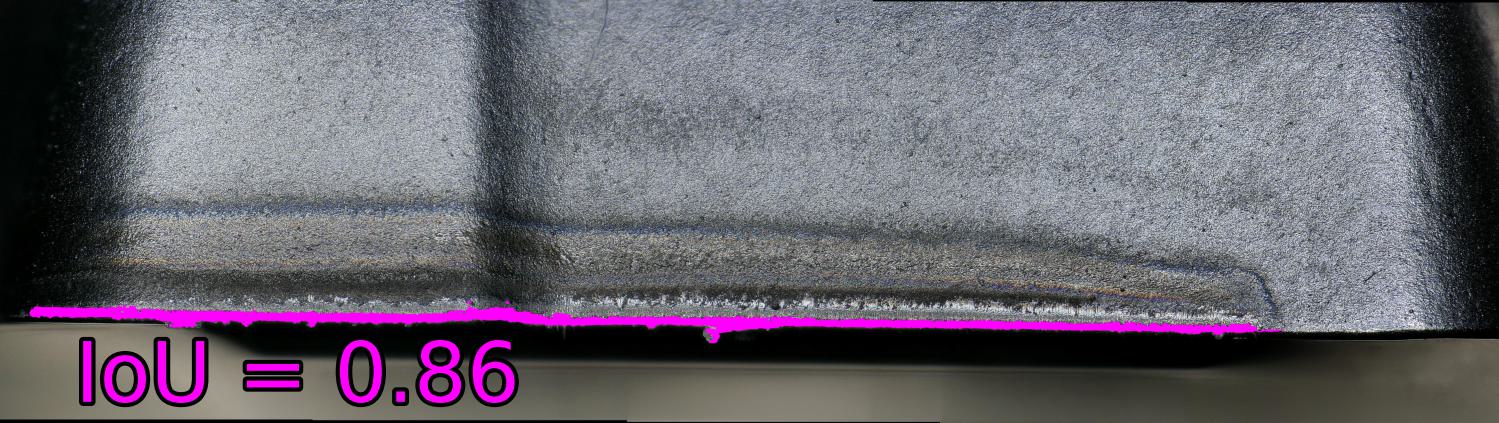}
        \hfill
        \includegraphics[width=.32\textwidth]{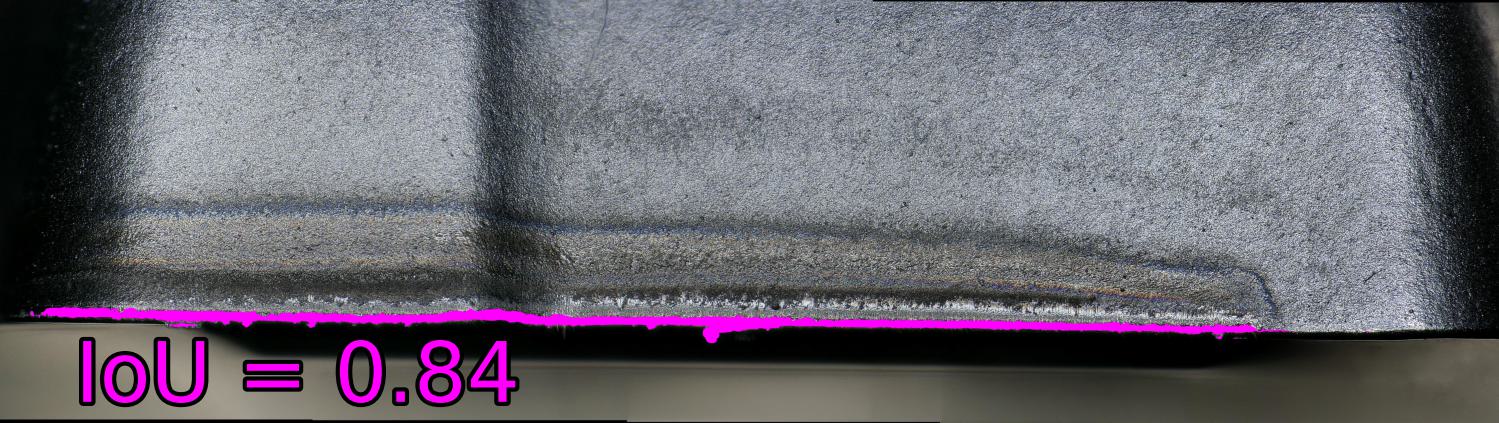}
    \vspace*{1mm}
    \end{subfigure}
    \begin{subfigure}[t]{\textwidth}
    \centering
        \includegraphics[width=.32\textwidth]{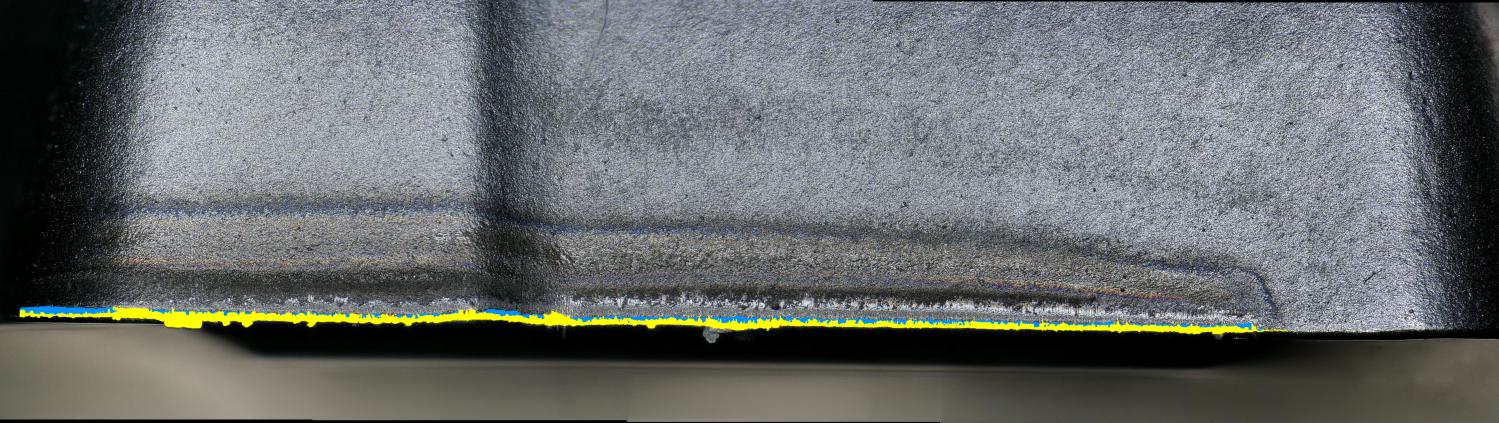}
        \hfill
        \includegraphics[width=.32\textwidth]{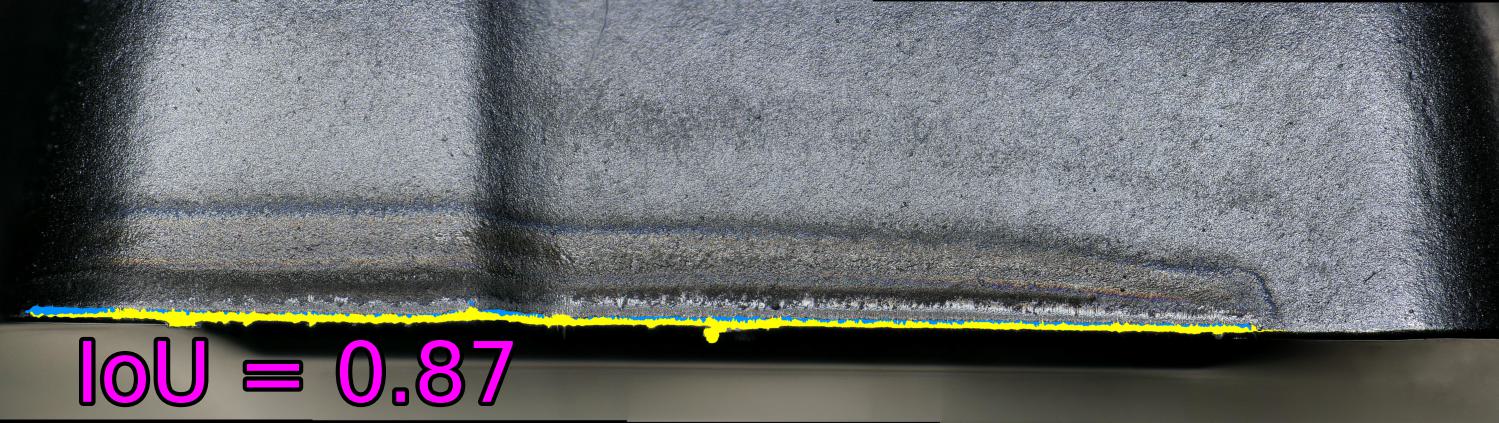}
        \hfill
        \includegraphics[width=.32\textwidth]{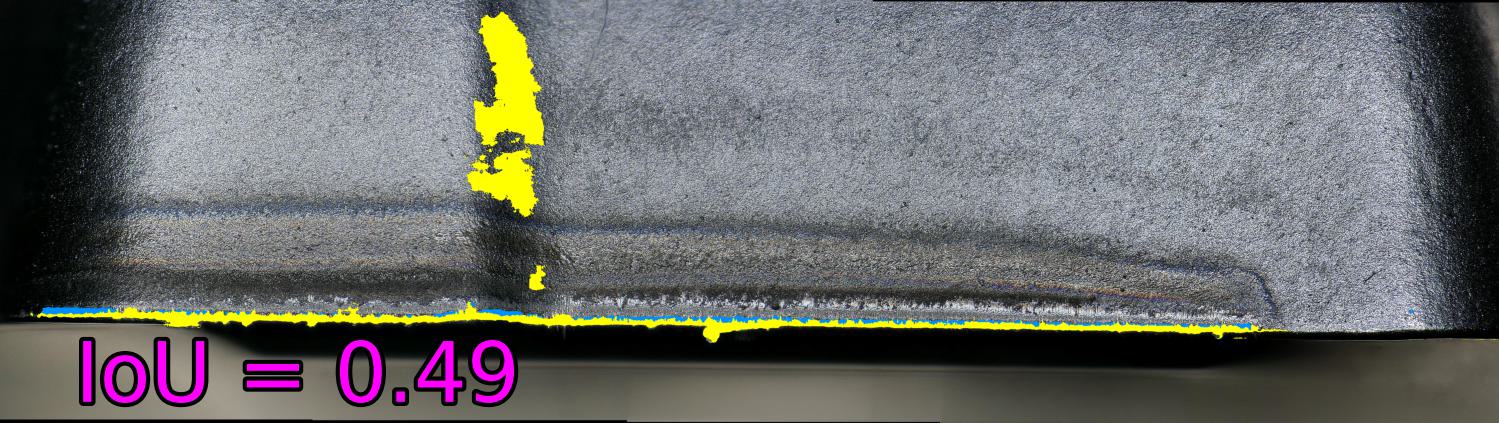}
    \vspace*{5mm}
    \end{subfigure}
    \begin{subfigure}[t]{\textwidth}
    \centering
        \includegraphics[width=.32\textwidth]{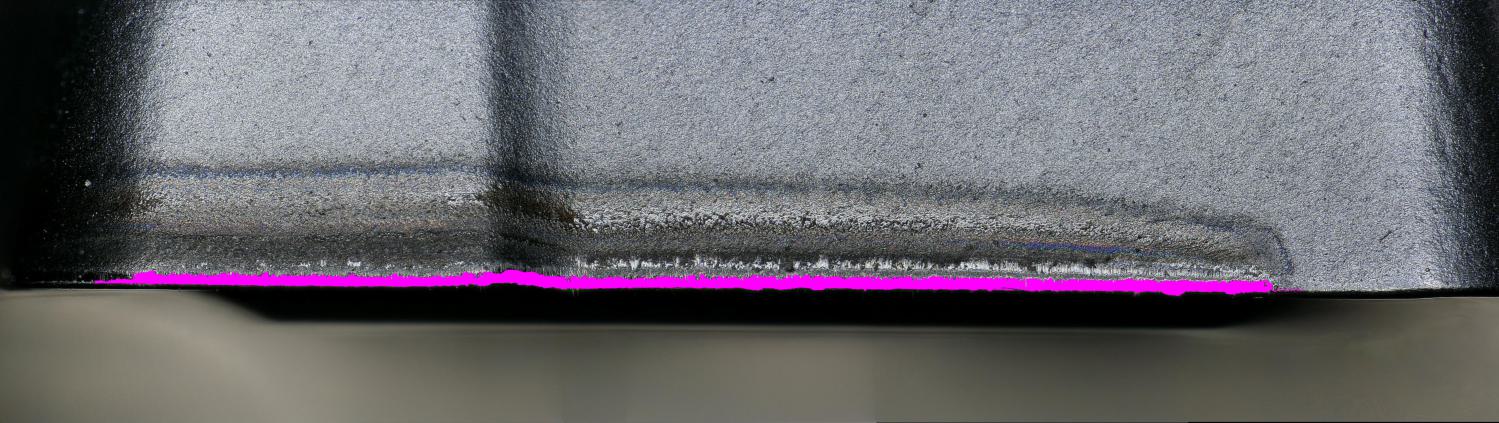}
        \hfill
        \includegraphics[width=.32\textwidth]{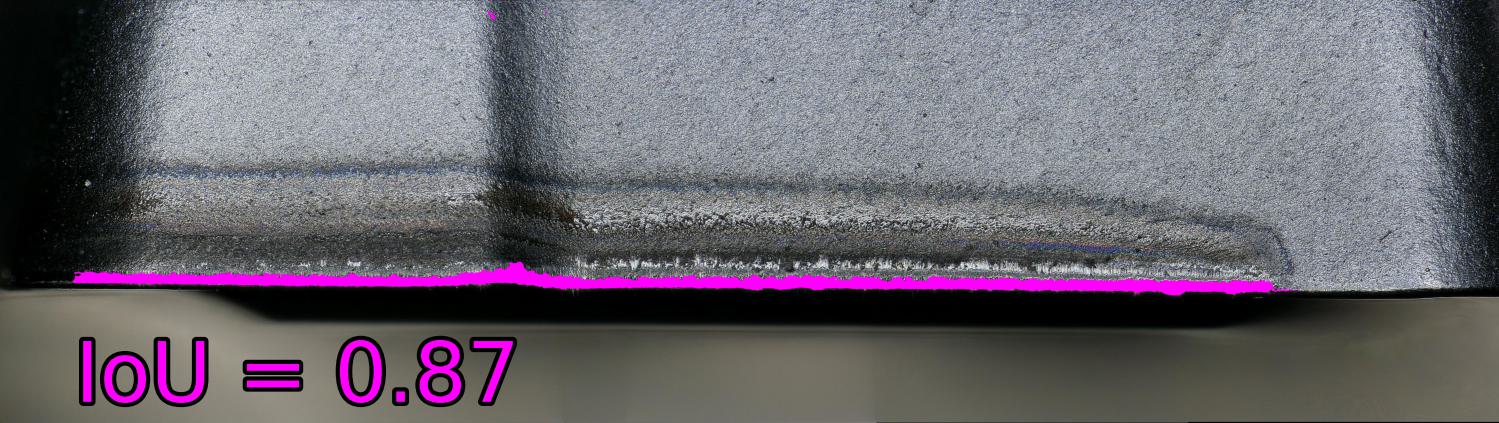}
        \hfill
        \includegraphics[width=.32\textwidth]{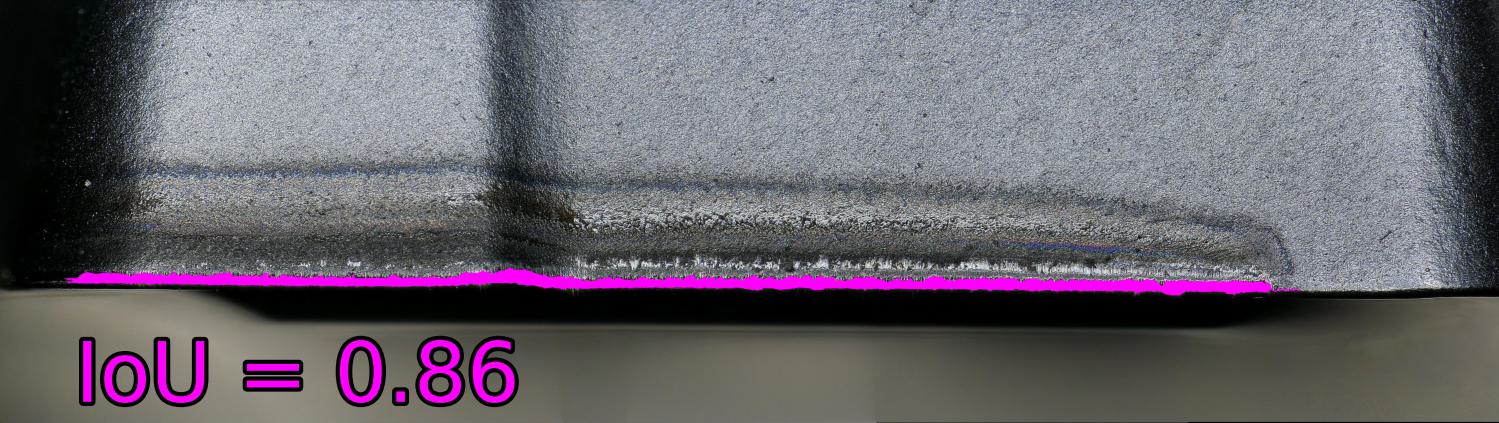}
    \vspace*{1mm}
    \end{subfigure}
    \begin{subfigure}[t]{\textwidth}
    \centering
        \includegraphics[width=.32\textwidth]{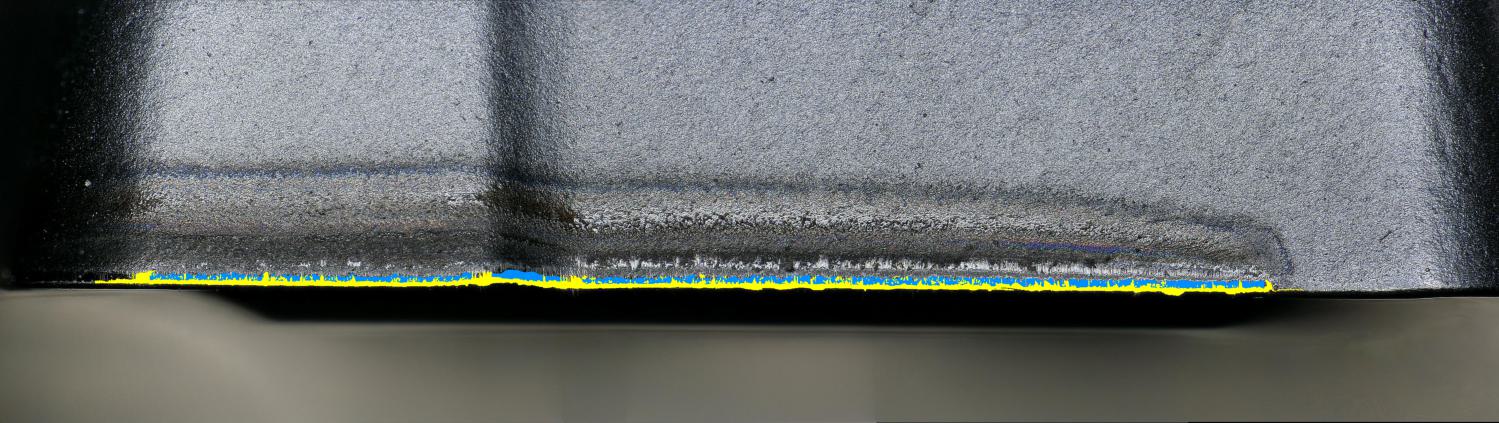}
        \hfill
        \includegraphics[width=.32\textwidth]{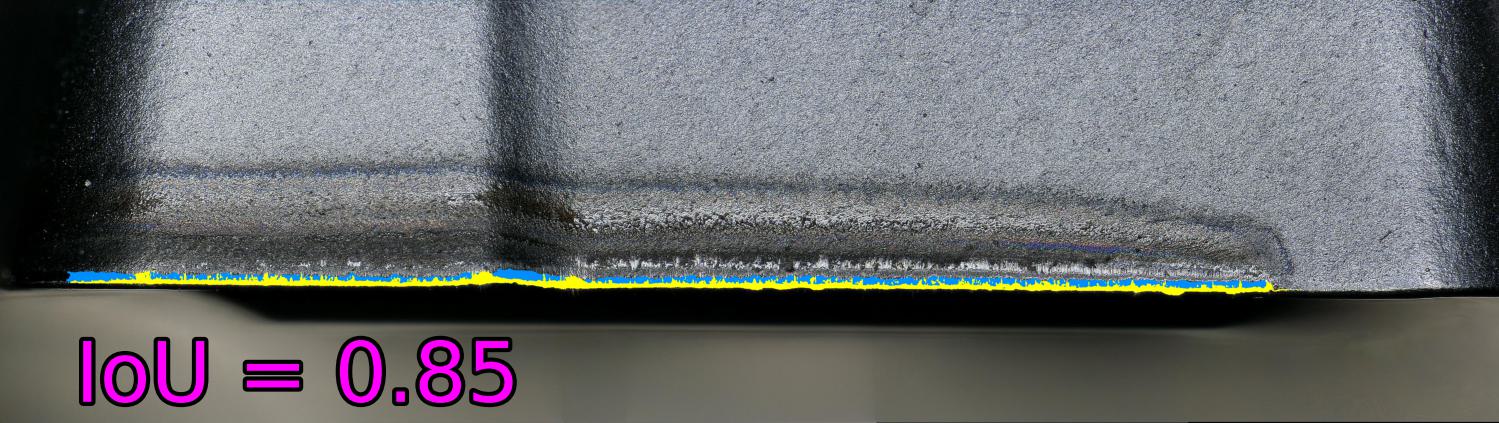}
        \hfill
        \includegraphics[width=.32\textwidth]{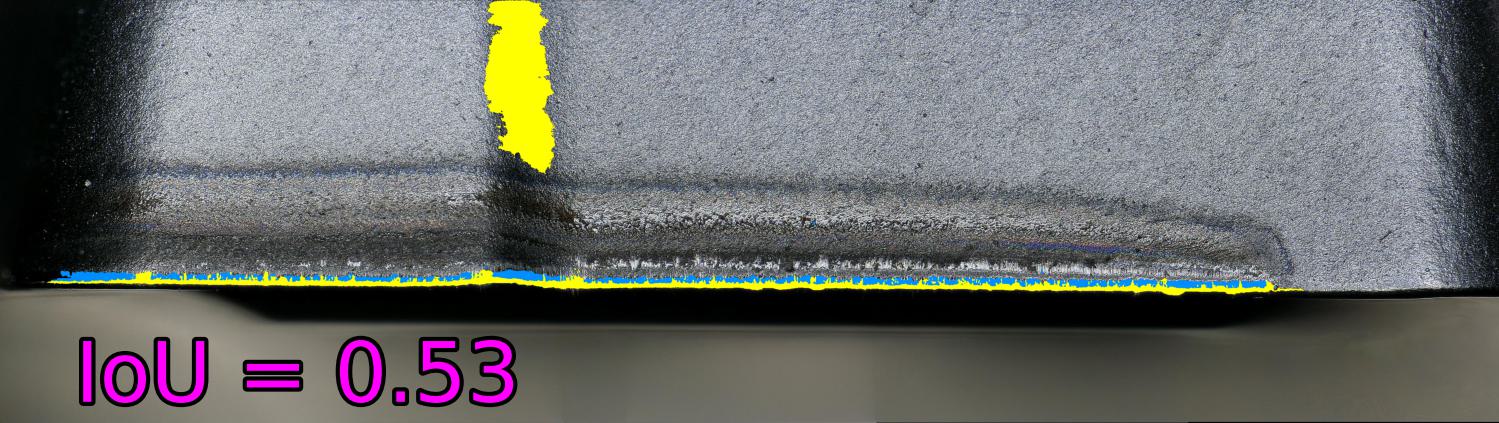}
    \vspace*{5mm}
    \end{subfigure}
    \begin{subfigure}[t]{\textwidth}
    \centering
        \includegraphics[width=.32\textwidth]{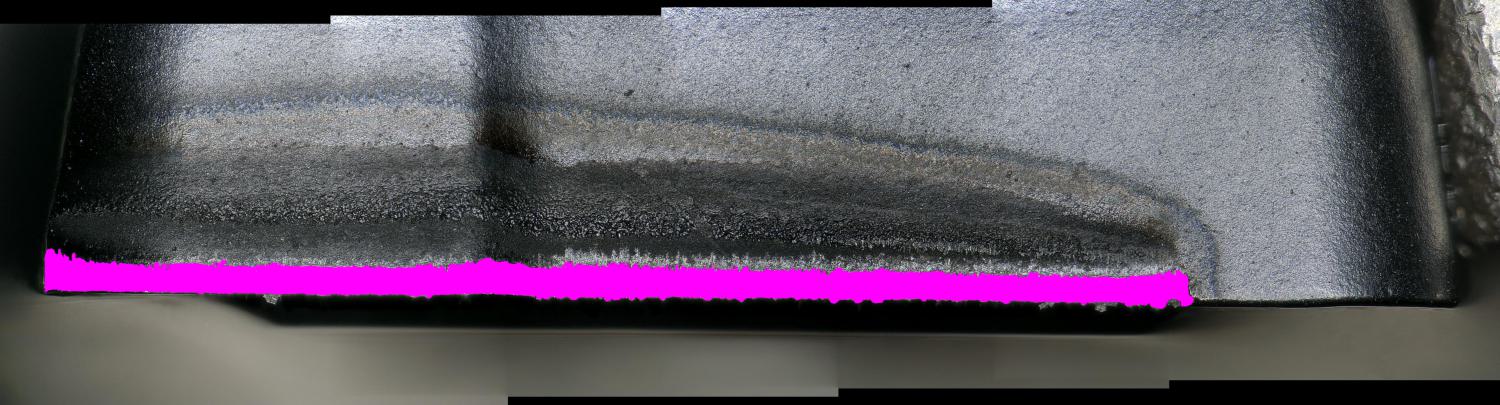}
        \hfill
        \includegraphics[width=.32\textwidth]{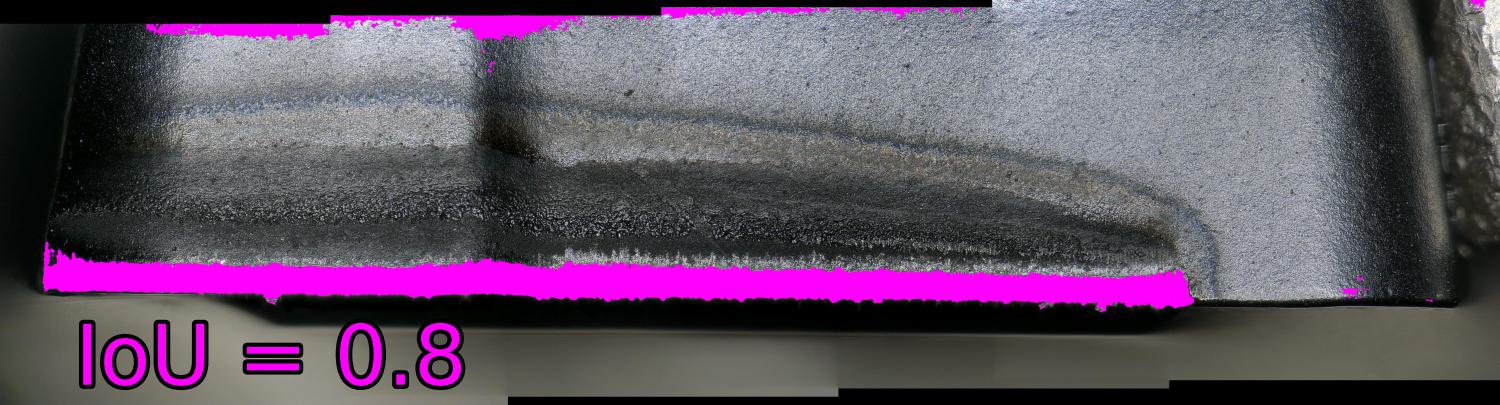}
        \hfill
        \includegraphics[width=.32\textwidth]{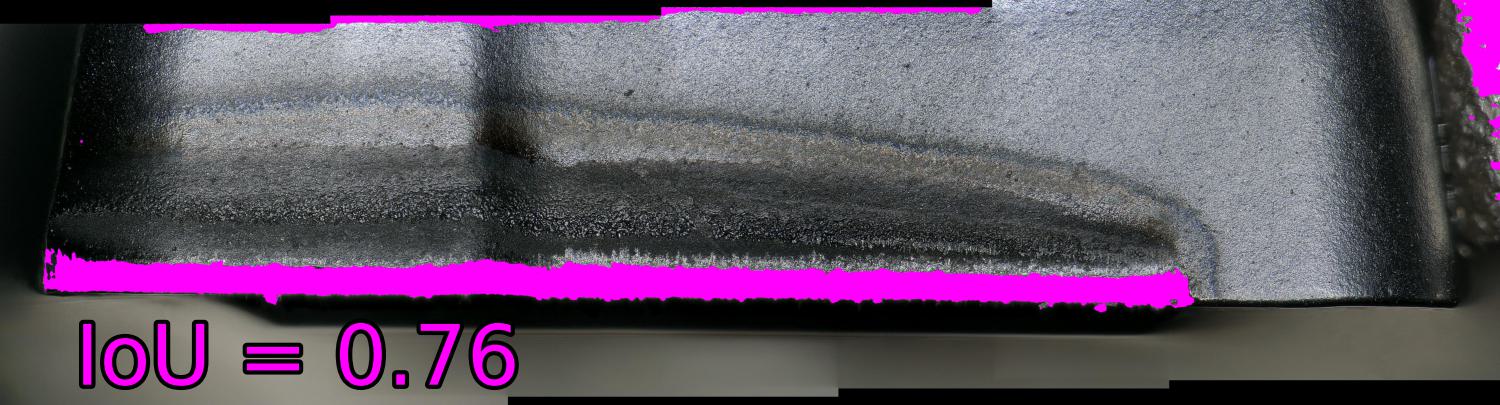}
    \vspace*{1mm}
    \end{subfigure}
    \begin{subfigure}[t]{\textwidth}
    \centering
        \includegraphics[width=.32\textwidth]{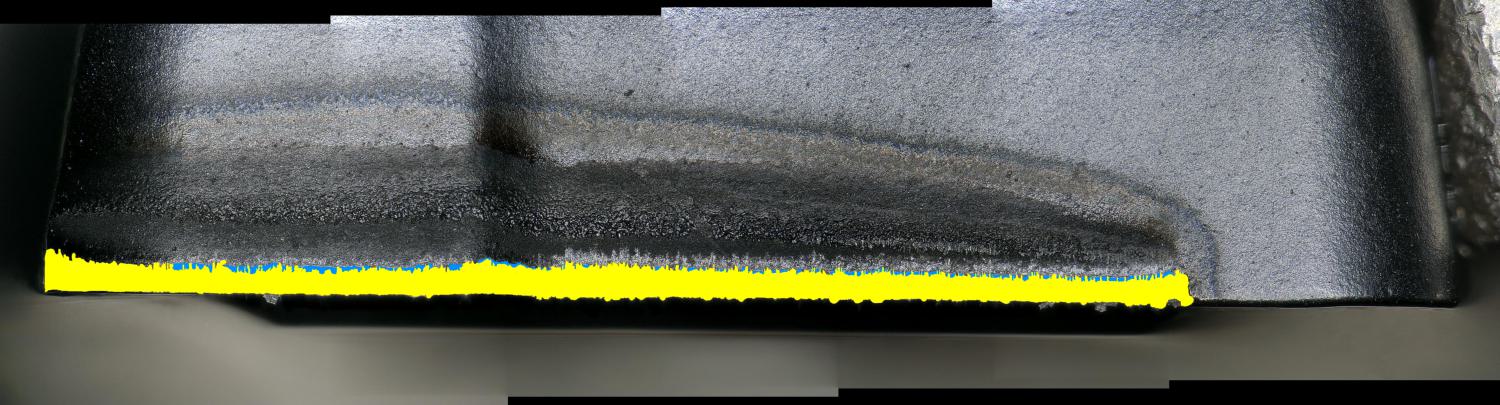}
        \hfill
        \includegraphics[width=.32\textwidth]{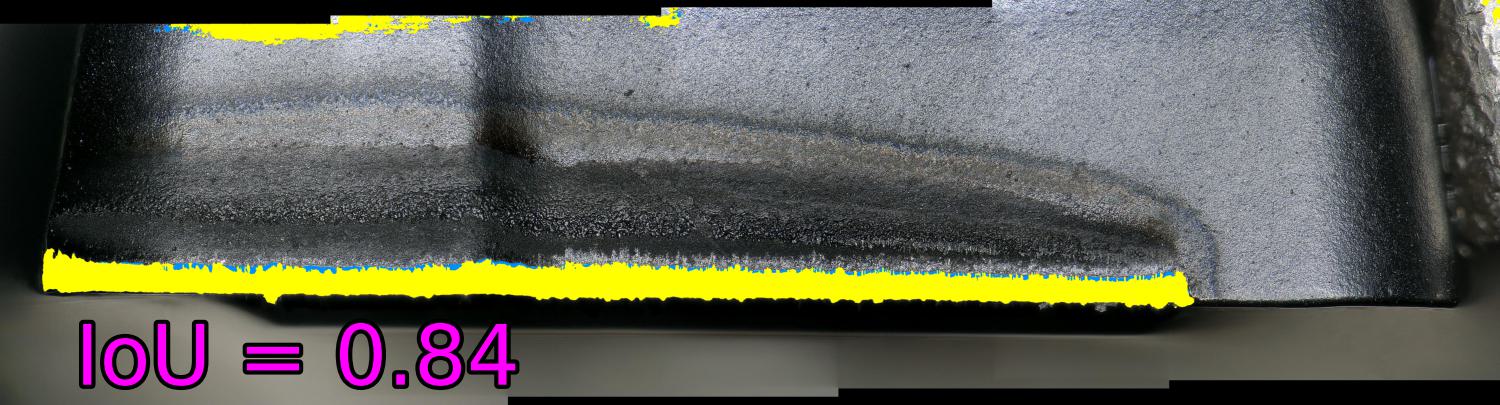}
        \hfill
        \includegraphics[width=.32\textwidth]{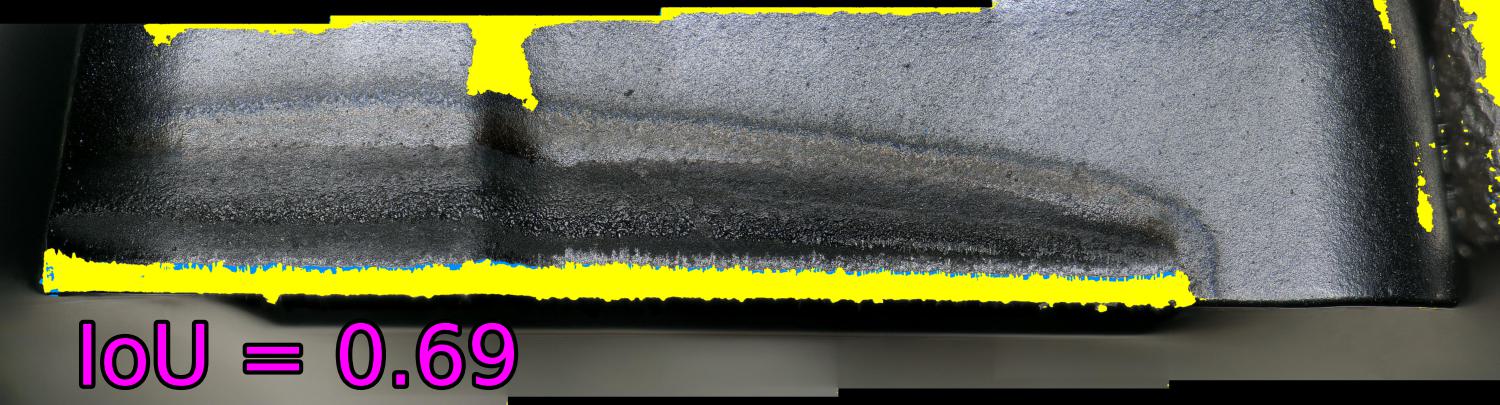}
    \vspace*{5mm}
    \end{subfigure}
    \begin{subfigure}[t]{\textwidth}
    \centering
        \includegraphics[width=.32\textwidth]{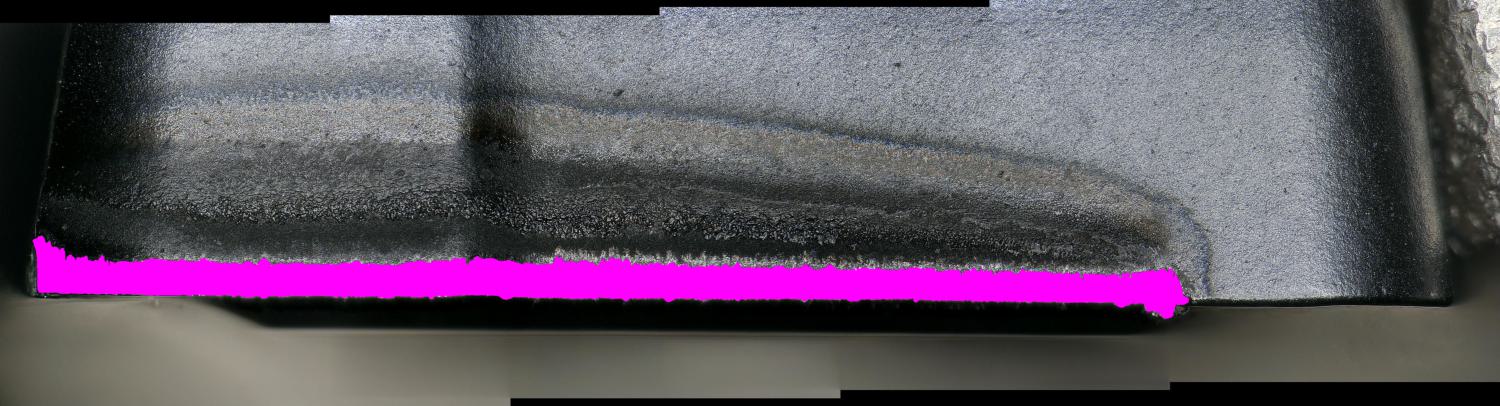}
        \hfill
        \includegraphics[width=.32\textwidth]{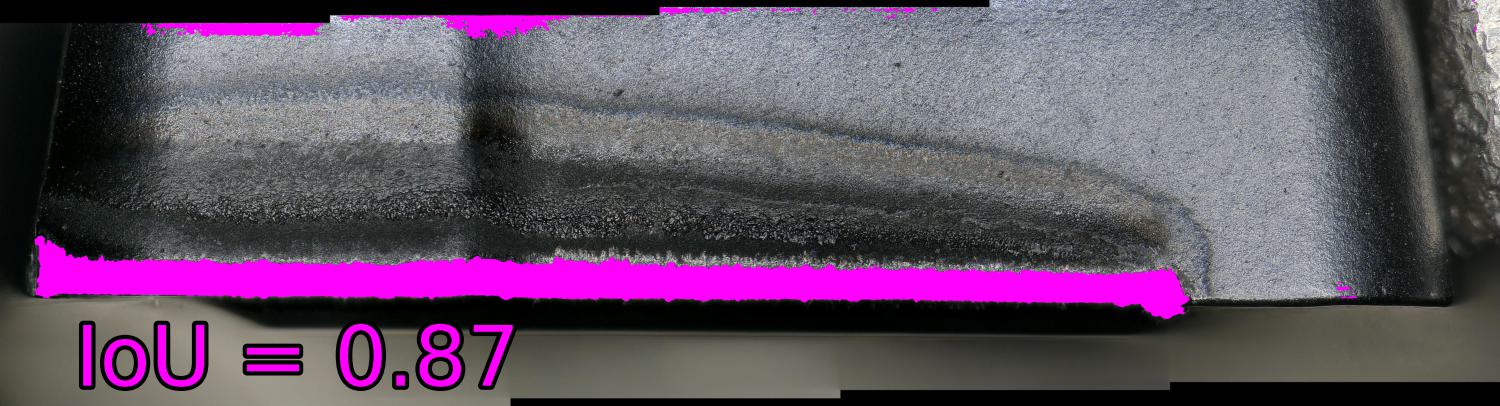}
        \hfill
        \includegraphics[width=.32\textwidth]{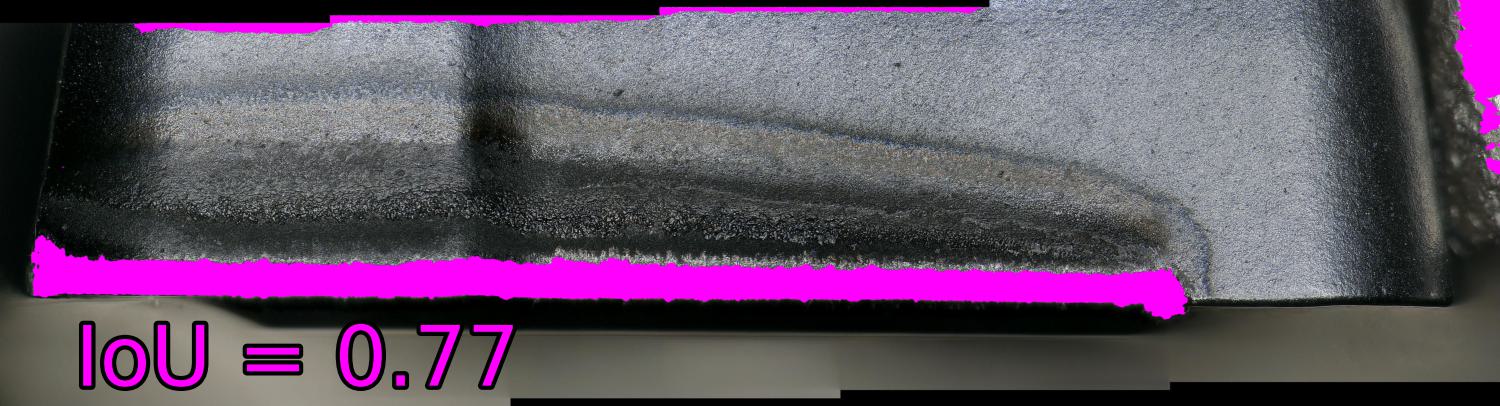}
    \vspace*{1mm}
    \end{subfigure}
    \begin{subfigure}[t]{\textwidth}
    \centering
        \includegraphics[width=.32\textwidth]{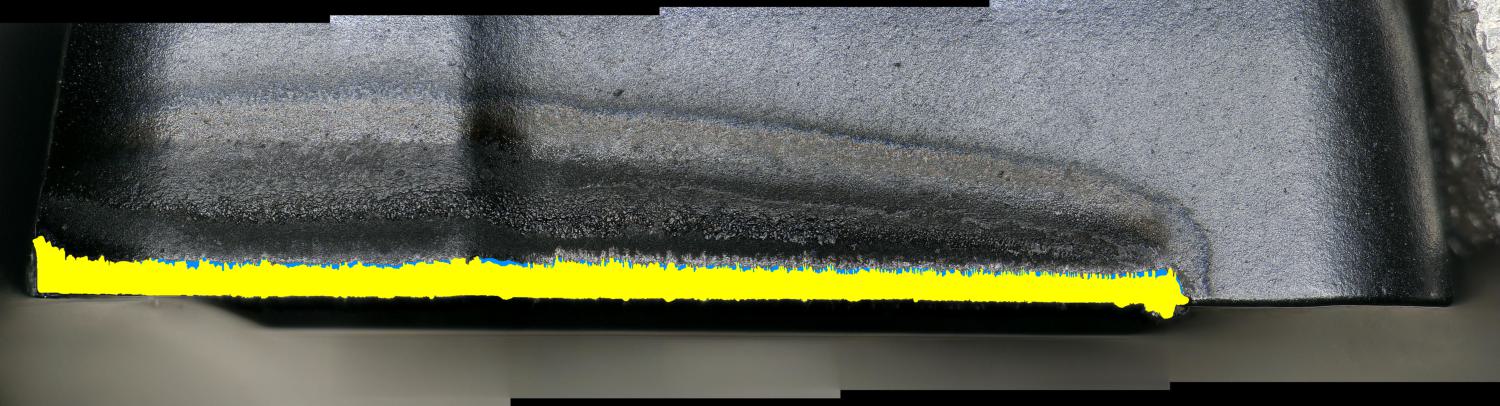}
        \hfill
        \includegraphics[width=.32\textwidth]{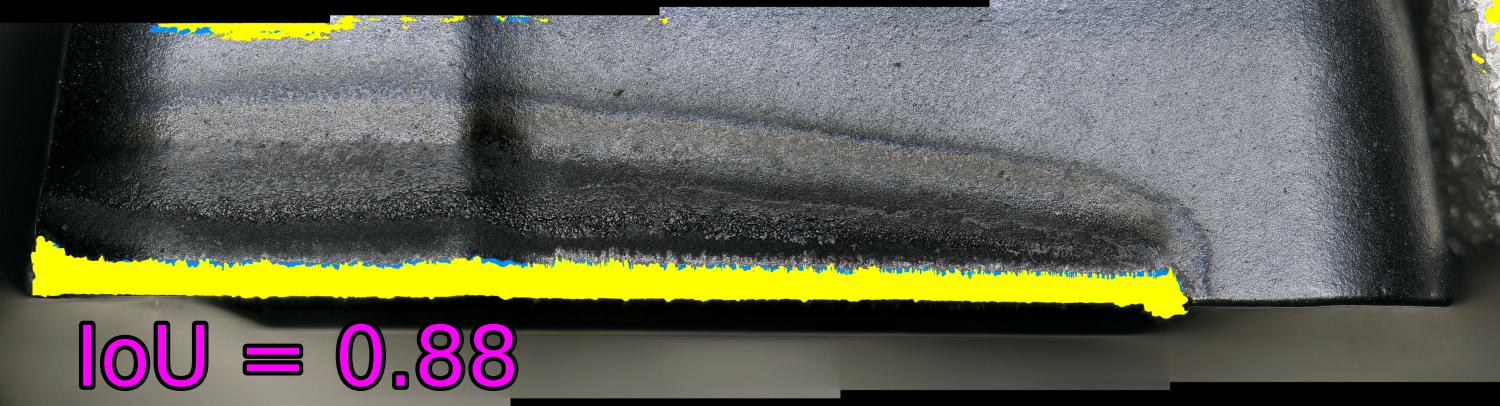}
        \hfill
        \includegraphics[width=.32\textwidth]{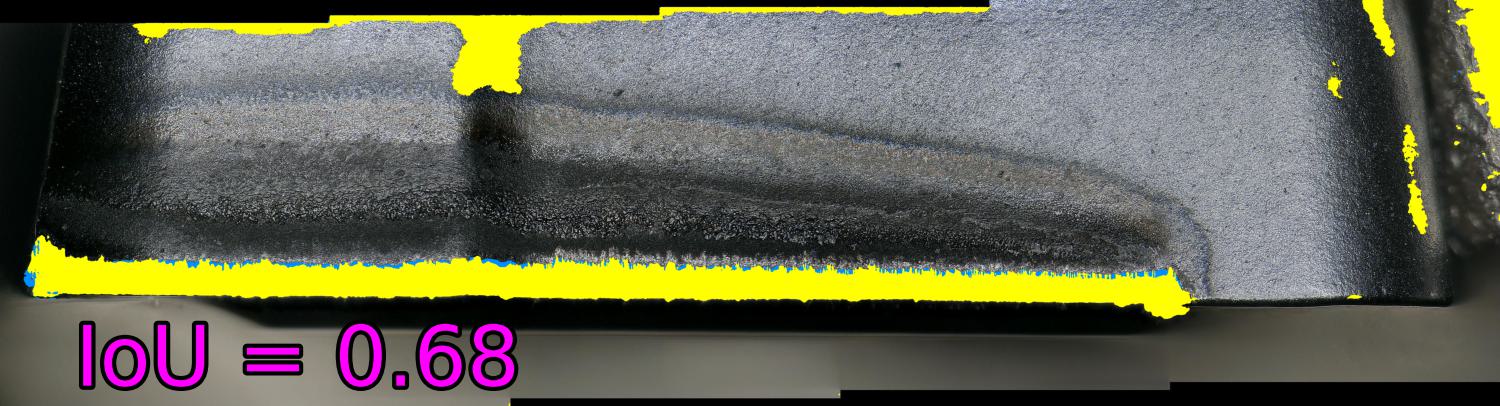}
    \vspace*{1mm}
    \end{subfigure}
\caption{Evaluation of the four test images, each arranged in blocks of two rows: In each block, the first image in the first row shows images with the binary ground truth mask (left), followed by the predictions of the binary model using tiles with edge length 512~px (middle), followed by images of tiles with edge length 256~px (right). The second row shows the same for the multiclass masks and multiclass models. While for the multi class models the predicted categories A and M are shown separately, the reported IoU values are computed using A and M as one joint class.}
\label{fig:pred}
\end{figure*}

The final predictions on the four test images as a whole are shown in Figure~\ref{fig:pred} for the four final models, whereby an overlap-tile strategy was used to get seamless predictions. For the first two images, the binary predictions are of very high quality, and both models perform very similar. The last two images show some problems: Firstly, there are some artefacts of image stitching from the microscopy process at the upper and lower edge of the images. At the upper edge, this leads to some false positive areas, since some black area adjoins the cutting insert surface. This pattern is similar to the cutting edge itself, where some wear is present in all the training tiles. Thus, it is not surprising that both models predict false positive wear areas. Furthermore, the two last images have an area with high reflection on the right side of the image, which is predicted as false positive by the model using smaller tile size only. This may be due to the smaller portion of background area in the tiles this model is trained on. 
The predictions using the multiclass models with tiles of edge length 512~px perform similar to the respective predictions of the binary models, with less false positives in the presence of the artefacts in the upper part of the last two images. In contrast, the multiclass model using tiles of edge length 256~px generate false positives not only at the artefacts in the upper part of the images, but also at the curved parts of the cutting insert, which present some differences in shadow and reflection.

\section{Conclusion \& Future Work}\label{sec:4}

In this work, we have investigated the feasibility of deploying a U-Net based image segmentation pipeline for wear detection on drilling tool devices. Thereby, a wide variety of different combinations of data augmentation strategies, tile sizes, and loss functions (binary and multiclass), as well as their impact on the model's performance, have been investigated. Among the tested configurations, the binary model with batch normalisation trained on data with moderate augmentation minimising the IoU-based loss performs best for tiles with edge length 512~px and 256~px, reaching IoUs of 0.886 and 0.904 respectively. Evaluation of the predictions on the full images via overlap-tile strategy indicates that those models that use larger tiles are more robust against high reflection areas as compared to models trained on smaller tiles.

While for non-augmented training data, the resulting model performance varies drastically with the choice of loss functions, the loss function does not have that large an impact when training with data that has been augmented moderately. The distinction of the wear types in the multiclass model seems to help in training. However, the weighting of the different classes within the loss functions was fixed, and more suitable weightings may improve multiclass models. 
Since data augmentation significantly affects the performance of the resulting model in this study, further exploration of data augmentation intensity and techniques is of great interest for future research. 

In summary, we have shown that U-Net based image segmentation strategies are fruitful approaches to tackle the challenging problem of wear detection for drilling tools. Using this direct approach to assess the wear of a tool can assist lab technicians to quickly detect and highlight wear in microscopy images. Furthermore, as our pipeline allows for extraction of the areas affected by wear, robust metrics can be conceived that are based upon, and directly derived from, the predicted segmentation mask, which could in turn help to formulate meaningful measures of wear in drilling processes.

\FloatBarrier



\section*{Acknowledgements}

The authors gratefully acknowledge the financial support under the scope of the COMET program within the K2 Center “Integrated Computational Material, Process and Product Engineering (IC-MPPE)” (Project No 886385). This program is supported by the Austrian Federal Ministries for Climate Action, Environment, Energy, Mobility, Innovation and Technology (BMK) and for Labour and Economy (BMAW), represented by the Austrian Research Promotion Agency (FFG), and the federal states of Styria, Upper Austria and Tyrol.

The Know-Center is funded within the Austrian COMET Program---Competence Centers for Excellent Technologies---under the auspices of the Austrian Federal Ministry for Climate Action, Environment, Energy, Mobility, Innovation and Technology (BMK), the Austrian Federal Ministry for Digital and Economic Affairs (BMDW) and by the State of Styria. COMET is managed by the Austrian Research Promotion Agency FFG.

We also want to thank Benno Spors, Marco Dieter, and Steffen Möss for creation of the digital microscopy images of the cutting inserts, and Manfred Mücke for his support in the project.

\bibliographystyle{elsarticle-num-names}   
\bibliography{main}

\end{document}